%% file: latex/main.tex
\documentclass[11pt]{article}

\usepackage[preprint]{acl}

\usepackage{times}
\usepackage{latexsym}
\usepackage{hyperref}
\usepackage{url}
\usepackage{bbm}
\usepackage[all]{tcolorbox}
\usepackage{multirow}
\usepackage{booktabs}
\usepackage{makecell}
\usepackage{pifont}
\usepackage[table]{xcolor}
\usepackage{graphicx}
\usepackage{subcaption}
\usepackage{geometry}
\usepackage{enumitem}
\usepackage{amsmath}
\usepackage{ulem}

\usepackage[T1]{fontenc}

\usepackage[utf8]{inputenc}

\usepackage{microtype}
\usepackage{inconsolata}

\usepackage{graphicx}

\definecolor{best}{HTML}{EBF8E1}
\definecolor{medium}{HTML}{FFF8E7}
\definecolor{worst}{HTML}{FBE4E5}
\definecolor{origin}{HTML}{FDB863}
\definecolor{int-bg}{HTML}{F2FCFD}
\definecolor{mxfp-bg}{HTML}{b5c6e0}
\newtcolorbox{mybox}[2][]
  {colback = black!5!white, colframe = black!75!black, fonttitle = \bfseries,
    colbacktitle = black!100!black, enhanced, before upper={\fontsize{10}{11}\obeyspaces\obeylines\selectfont}, fontupper=\selectfont,
    attach boxed title to top left={yshift=-2.2mm,xshift=2mm},
    title=#2,#1}
\usepackage{tikz}

\title{Benchmarking Post-Training Quantization of Large Language Models under Microscaling Floating Point Formats}

\author{
 \textbf{Manyi Zhang$^{*}$}\ \ \
 \textbf{Ji-Fu Li$^{*}$} \ \ \
 \textbf{Zhongao Sun}\thanks{Equal contribution. \ $^{\dag}$Corresponding author.}\\
 \textbf{Haoli Bai} \ \ \
 \textbf{Hui-Ling Zhen} \ \ \
 \textbf{Zhenhua Dong} \ \ \
 \textbf{Xianzhi Yu$^{\dag}$}
 \\
 \\
  Huawei Technologies
\\
 \small{
   \{zhangmanyi6, yuxianzhi\}@huawei.com}
    }

\begin{document}
\maketitle
\begin{abstract}
Microscaling Floating-Point (MXFP) has emerged as a promising low-precision format for large language models~(LLMs). Despite various post-training quantization (PTQ) algorithms being proposed, they mostly focus on integer quantization, while their applicability and behavior under MXFP formats remain largely unexplored. To address this gap, this work conducts a systematic investigation of PTQ under MXFP formats, encompassing over 7 PTQ algorithms, 15 evaluation benchmarks, and 3 LLM families. The key findings include: 1) MXFP8 consistently achieves near-lossless performance, while MXFP4 introduces substantial accuracy degradation and remains challenging; 2) PTQ effectiveness under MXFP depends strongly on format compatibility, with some algorithmic paradigms being consistently more effective than others; 3) PTQ performance exhibits highly consistent trends across model families and modalities, in particular, quantization sensitivity is dominated by the language model rather than the vision encoder in multimodal LLMs; 4) The scaling factor of quantization is a critical error source in MXFP4, and a simple pre-scale optimization strategy can significantly mitigate its impact. Together, these results provide practical guidance on adapting existing PTQ methods to MXFP quantization.
\end{abstract}

\input{latex/introduction}
\input{latex/preliminary}
\input{latex/experiments}
\input{latex/conclusion}

\bibliography{custom}
\clearpage
\appendix

\input{latex/appendix}

\end{document}

%% file: latex/introduction.tex
\section{Introduction}
The ever-growing scale of large language models (LLMs) poses a significant deployment challenge due to prohibitive demands for memory and computational resources~\citep{yuan2024llm, dantas2025review}. Quantization has therefore emerged as an indispensable technique to mitigate these issues by reducing the
numerical precision of weights or activations~\citep{nagel2021white, kurtic2025give, huang2024good, guo2025deepseek,tang2025pangu}. Among various low-precision representations, Microscaling Floating-Point (MXFP) formats have attracted increasing attention~\citep{shao2025block}. As a block-level number format, MXFP better preserves the dynamic range of full-precision models while benefiting from growing hardware support~\citep{amd2025cdna4, choquette2023nvidia, tirumala2024nvidia}.

In this context, post-training quantization (PTQ) provides a practical and training-free route to compress pre-trained models, and a rich line of PTQ techniques has been developed to alleviate the accuracy drop under low precision. However, existing PTQ research and design choices have been predominantly tailored to integer (INT) formats~\citep{frantar2022gptq, shao2023omniquant, ma2024affinequant, lin2024duquant, dettmers2022gpt3, li2024svdquant, wang2023bitnet}. This leaves the effectiveness and potentially the distinct failure modes of PTQ on MXFP formats largely underexplored.

\begin{figure}[!t]
    \centering
    \includegraphics[width=0.98\linewidth]{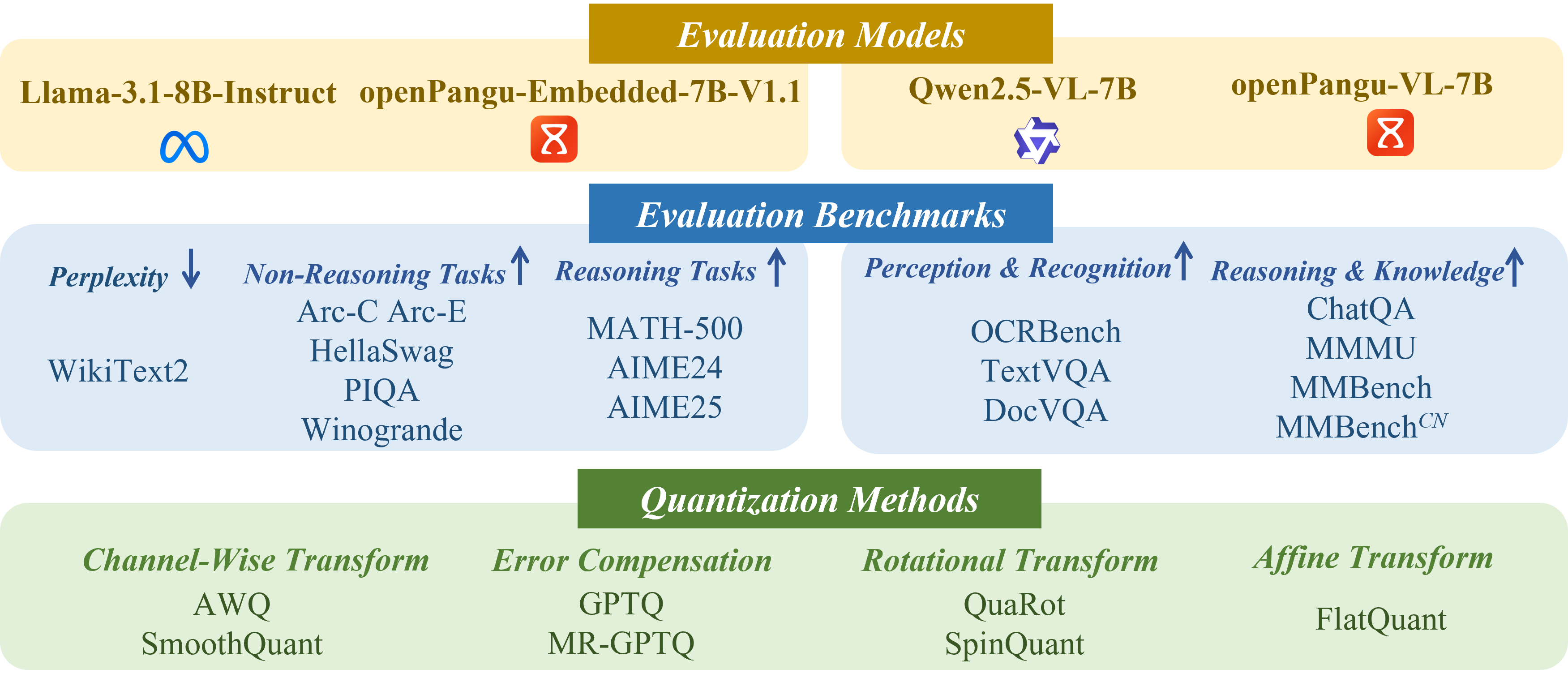}
    \vspace{-0.175cm}
    \caption{The overview of our empirical evaluations.}
    \vspace{-0.3cm}
    \label{fig:overview}
\end{figure}

In this paper, we provide a comprehensive empirical study of various PTQ methods for MXFP formats, including MXFP8 and MXFP4, as outlined in Figure~\ref{fig:overview}. We systematically categorize existing PTQ methods into four classes: (i) channel-wise transformation, (ii) error compensation, (iii) rotational transformation, and (iv) affine transformation. Our evaluations cover weight-only, weight-activation, and KV cache quantization across multiple bit-width configurations, and span both reasoning and non-reasoning benchmarks. The studied models encompass two different LLM families (\textit{i.e.}, Llama3.1-8b-Instruct~\citep{grattafiori2024llama} and openPangu-Embedded-7B-v1.1~\citep{chen2025pangu}) and multimodal LLMs~(MLLM) (\textit{i.e.}, Qwen2.5-VL-7B~\citep{Qwen2VL} and openPangu-VL-7B~\citep{openPangu-VL}). Based on extensive experiments, we distill several key findings, summarized as follows:

\begin{itemize}[leftmargin=20pt]
    \item[1.] \textbf{Lossless Quantization} (\S\ref{sec:rq1}): 8-bit weight-activation MXFP quantization is lossless across tasks and modalities. In contrast, 4-bit weight or activation quantization under MXFP results in non-negligible accuracy degradation and remains an open challenge.
    \item[2.] \textbf{Quantization Algorithms} (\S\ref{sec:rq2}): Error compensation and affine transformation methods are more compatible with MXFP quantization, particularly at low bit-widths, and their combination yields stronger performance. Notably, RTN remains a competitive baseline, indicating that MXFP demands quantization methods tailored to its specific scheme.
    \item[3.] \textbf{Impact of Model Family and Modalities} (\S\ref{sec:rq3}): PTQ under MXFP exhibits consistent effectiveness across different model families and modalities. 
    For MLLMs, quantization sensitivity is dominated by the LLM component rather than the vision encoder, favoring mixed-precision designs that preserve higher precision in the LLM. 
    Visual tokens are comparatively robust under MXFP, with reduced bit-width incurring negligible accuracy loss.
    \item[4.] \textbf{Quantization Components} (\S\ref{sec:rq4}): The quantization error introduced by the scaling factors is noticeable. Pre-scale operation, a practically effective optimization strategy for MXFP4, is recommended.
\end{itemize}

Our empirical study and findings show that MXFP is not merely a drop-in replacement for existing low-precision formats, but also a distinct numerical regime that calls for format-aware quantization design. We hope this work serves as a reference point for future research on MXFP-centric PTQ methods, and encourages the community to move beyond integer-oriented assumptions when developing low-precision algorithms for large-scale foundation models.

%% file: latex/preliminary.tex
\section{Preliminary}

\subsection{Low-Bit Integer (INT) and Floating-Point (FP) Quantization}
Quantization maps a tensor $\mathbf{W}$ in high-precision source-format to a lower bit-width, such as low-bit INT and FP formats. For integer quantization, we define: 
\begin{equation}
\mathbf{W}_\text{q}:=\texttt{clip}\left(\left\lfloor\mathbf{W}/s\right\rceil,Q_{\min}, Q_{\max}\right)\cdot s,
\end{equation}
where $\texttt{clip}(\lfloor\mathbf{W}/s\rceil, Q_{\min}, Q_{\max})$ truncates the associated values inside the minimal $Q_{\min}$ and maximal
$Q_{\max}$, and $s$ is the scaling factor that normalizes $\mathbf{W}$ to the target integer range. More complex than integers, floating-point numbers are encoded using three components~\citep{markstein2008new}: a sign bit ($S$), an exponent ($E$), and a mantissa ($M$). We denote a format as $\text{E}a\text{M}b$, where $a$ and $b$ are the exponent and mantissa bit numbers. 
Floating-point quantization is defined as:
\begin{equation}
\mathbf{W}_\text{q}:=\texttt{nearest}\left(\left\lfloor\mathbf{W}/s\right\rceil, \mathbbm{C}_{\text{FP}}\right)\cdot s,
\end{equation}
where $\mathbbm{C}_{\text{FP}}$ is the set of representable low-bit floating-point values related to $S$, $E$, and $M$~\citep{chen2025int}, and $\texttt{nearest}(\cdot,\mathbbm{C}_{\text{FP}})$ maps normalized values to the nearest element of $\mathbbm{C}_{\text{FP}}$. 

\subsection{Microscaling Floating Point (MXFP) Quantization}
The MXFP, proposed by OCP~\citep{rouhani2023microscaling}, is a family of quantized floating-point formats that utilize block quantization. An MX-format is specified by the block size 32 and uses a shared UE8M0 data-type for each block. The MXFP8 format has E4M3 and E5M2 variants, and MXFP4 is E2M1. Here, we adopt E4M3 for MXFP8, as a larger mantissa width is more crucial for the performance of fine-grained quantization~\citep{mishra2025recipes, chen2025oscillation}.

\subsection{Post-training Quantization Methods}\label{sec:ptq_methods}
We consider quantization under PTQ. Here, we divide PTQ methods into four categories: channel-wise transformation, error compensation, rotational transformation, and affine transformation.  For each category, we evaluate representative algorithms that embody its core principle.

\paragraph{Channel-Wise Transformation.} Generally, this kind of method aims to reduce quantization error by adaptively adjusting the numerical ranges of activation and weights along individual channels. We here evaluate two prominent methods, which are SmoothQuant~\citep{xiao2023smoothquant} and AWQ~\citep{lin2024awq}. SmoothQuant proposes per-channel scaling to migrate quantization difficulty from activations to weights. AWQ is developed to enable efficient quantization of LLMs while preserving the precision of the most critical weights.

\paragraph{Error Compensation.} These methods aim to mitigate quantization-induced discrepancies by explicitly modeling and compensating for the quantization error. In this work, we consider GPTQ~\citep{frantar2022gptq} and MR-GPTQ~\citep{egiazarian2025bridging}. GPTQ performs layer-wise quantization and leverages inverse-Hessian information to update weights, thereby reducing accuracy degradation. MR-GPTQ extends GPTQ to better match the characteristics of FP4 quantization by incorporating block-wise Hadamard transforms and format-specific optimizations.

\paragraph{Rotational Transformation.} This class of methods utilizes pre-quantized orthogonal transformations to the weight and activations. These transformations can reconstruct the data distribution to mitigate the impact of extreme outliers on activations. We assess QuaRot~\citep{ashkboos2024quarot}, which uses random orthogonal rotations, and SpinQuant~\citep{liu2024spinquant}, which employs learnable rotations optimized during calibration. 

\paragraph{Affine Transformation.} These methods improve low-bit model compression by applying learnable, rescaling transformations to weights and activations before quantization, explicitly redistributing numerical magnitudes across dimensions. We include FlatQuant~\citep{sun2024flatquant}. To achieve flatter distributions of weights and activations, FlatQuant identifies layer-wise optimal affine transformations by employing a lightweight, block-wise training strategy over the calibration stage.

%% file: latex/experiments.tex
\section{Evaluations}
\subsection{Setup}
\paragraph{Quantization Settings.} We study several MXFP-based quantization configurations. (1) \textit{Weight-Only Quantization.} Only the weight tensors of linear layers are quantized using MXFP formats, while activations remain in full precision. (2) \textit{Weight-Activation Quantization.} Weights and input activation tensors are quantized using MXFP, enabling fully quantized matmul operations. (3) \textit{KV Cache Quantization.} The key and value tensors in attention blocks are quantized to reduce the memory footprint during inference. For clarity, we denote each configuration using the format $\text{W}\{\text{bits}\}\text{A}\{\text{bits}\}[\text{KV}\{\text{bits}\}]$. For example, W4A8 indicates quantizing weights to 4-bit and activations to 8-bit.

\paragraph{Evaluation Benchmarks.} We assess quantized models on the following benchmarks. (1) Language modeling quality via perplexity (PPL) on WikiText2~\citep{merity2016pointer}. (2) Non-reasoning tasks (zero-shot), including PIQA~\citep{bisk2020piqa}, Winogrande~\citep{sakaguchi2021winogrande}, Hellaswag~\citep{zellers2019hellaswag}, ARC-Easy~\citep{clark2018think}, and ARC-Challenge~\citep{clark2018think}. (3) Reasoning benchmarks, such as MATH-500~\citep{lightman2023let}, AIME24 and AIME25. (4) Multimodal benchmarks, including OCRBench~\citep{liu2024ocrbench}, MMBench~\citep{liu2024mmbench}, $\text{MMBench}^{CN}$~\citep{zhang2025lmms}, TextVQA~\citep{singh2019towards}, ChartQA~\citep{masry2022chartqa}, MME~\citep{fu2025mme}, and MMMU~\citep{yue2024mmmu}. To simulate the quantization with MXFP format, we use the microxcaling library\footnote{https://github.com/microsoft/microxcaling} for all experiments. Details of the evaluation are provided in Appendix \ref{appendix:evaluation}.

\paragraph{Evaluation Models.} We evaluate two popular LLMs, including Llama-3.1-8B-Instruct~\citep{grattafiori2024llama} and openPangu-Embedded-7B-V1.1~\citep{chen2025pangu}. Llama-3.1-8B-Instruct is an instruction-tuned LLM based on the Llama-3 architecture, offering strong multilingual and reasoning capabilities. openPangu-Embedded-7B-V1.1 is an efficient reasoning-focused LLM featuring a dual-system (fast/slow) inference capability. To further investigate the performance of MXFP quantization on the MLLM, we additionally evaluate the Qwen2.5-VL-7B~\citep{Qwen2VL} and openPangu-VL-7B models~\citep{openPangu-VL}.

We explore the following questions in the subsequent sections.

\begin{tcolorbox}
[colback=gray!10,colframe=gray!35!black, left=1ex]
\begin{description}    
    \item[RQ1] (\S\ref{sec:rq1}): What is the impact of different MXFP formats on model accuracy?
    \item[RQ2] (\S\ref{sec:rq2}): How do various post-training quantization methods perform with MXFP?
    \item[RQ3] (\S\ref{sec:rq3}): How do different model families and modalities affect quantization?
    \item[RQ4] (\S\ref{sec:rq4}): How do quantization design choices impact MXFP4 performance?
\end{description}
\end{tcolorbox}

\subsection{Performance on Different MXFP Quantization Settings~(RQ1)}
\label{sec:rq1}
\input{tables/llama8b}
\input{tables/pangu7b}
\input{tables/qwen_vl}
Following prior work~\citep{liu2025quantization}, we categorize post-quantization performance degradation into three regimes based on average accuracy recovery rate relative to BF16: \textcolor[HTML]{009900}{lossless ($\leq$1\%)}, \textcolor{origin}{fair (1\%–3\%)}, and \textcolor{red}{risky ($\geq$3\%)}. The overall results achieved by Llama-3.1-8B-Instruct, openPangu-Embedded-7B-V1.1, and Qwen2.5-VL-7B are summarized in Tables~\ref{tab:llama_nonreasoning}, \ref{tab:pangu_nonreasoning}, and \ref{tab:qwenvl}. We also provide the results of reasoning tasks in Figure~\ref{fig:reasoning} (details in Appendix~\ref{appendix:reasoning}). Note that the results across W8A8, W4A8, and W4A4 settings reveal a clear spectrum of quantization difficulty, which is largely dictated by the bit width of weights and activations. Below, we discuss these results.

\begin{figure}[!t]
    \centering
    \includegraphics[width=0.8\linewidth]{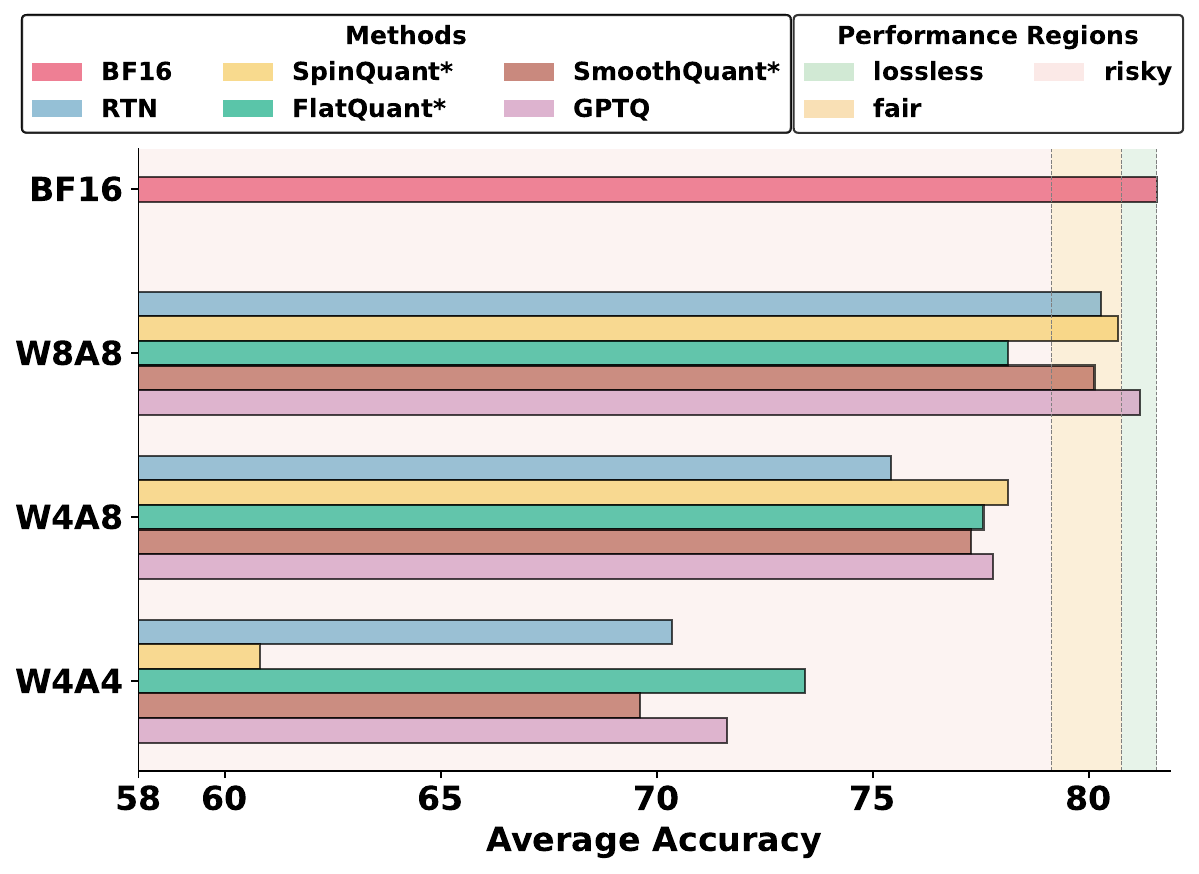}
    \vspace{-0.25cm}
    \caption{Average accuracy of Reasoning tasks on openPangu-Embedded-7B-V1.1.}
    \vspace{-0.25cm}
    \label{fig:reasoning}
\end{figure}

\paragraph{W8A8 is generally safe for deployment.} According to Tables \ref{tab:llama_nonreasoning}, \ref{tab:pangu_nonreasoning}, and \ref{tab:qwenvl}, the majority of PTQ methods consistently achieve lossless performance regardless of their internal design. This suggests that 8-bit quantization of both weights and activations is largely benign for modern LLMs and MLLMs, and can be safely deployed without extensive calibration or architectural adaptation.

\paragraph{W4A8 should be approached with caution.} We find that the transition to W4A8 with RTN incurs noticeable accuracy degradation, marking a critical inflection point. However, with PTQ algorithms, performance loss can be mitigated. As seen in Table~\ref{tab:pangu_nonreasoning}, the accuracy recovery rate is improved from 95.44\% to 98.60\% after optimization on openPangu-Embedded-7B-V1.1 and achieves near-lossless on Llama-3.1-8B-Instruct. On reasoning tasks, the accuracy improves significantly, but still falls into the risky region. These results indicate that while 4-bit weights challenge the model’s representational capacity, algorithmic refinements can enable W4A8 to be more viable, especially for non-reasoning tasks.

\paragraph{W4A4 is highly risky.} The most challenging setting is W4A4, where weights and activations are aggressively quantized. Performance degradation becomes more severe and widespread: accuracy recovery rate falls to 87.25\%–96.79\%,  86.37\%–95.28\%, and 92.72\%–97.36\% for Llama-3.1-8B-Instruct, openPangu-Embedded-7B-V1.1, and Qwen2.5-VL-7B, respectively, entering the risky regime for nearly all methods. Interestingly, the performance gap across different PTQ methods is markedly amplified under this extreme compression. This suggests that when quantization noise exceeds a critical threshold, differences in how methods handle outlier distributions, dynamic range alignment, or activation sensitivity become decisive.

\begin{mybox}[
  colback=gray!10,
  title=\textbf{Takeaways~1},
  boxrule=0.8pt,
  arc=2mm
]

W8A8 is consistently lossless across models and benchmarks, and is robust to the choice of PTQ methods. Bridging the performance gap for 4-bit weight or activation quantization~(\textit{cf.}, W4A8 and W4A4) under MXFP remains an open challenge.
\end{mybox}

\subsection{Comparison of PTQ Methods~(RQ2)}
\label{sec:rq2}
We evaluate diverse PTQ methods under MXFP (see the method descriptions in Section~\ref{sec:ptq_methods}). The main paper covers bit-width configurations of W8A8, W4A8, and W4A4, as shown in Tables~ \ref{tab:llama_nonreasoning}, \ref{tab:pangu_nonreasoning}, and \ref{tab:qwenvl}. Results on openPangu-VL-7B and more bit-width configurations (including W4A16 and W4A8KV8) can be found in Appendix \ref{appendix:pangu7bvl} and \ref{appendix:more_senario}, respectively. Below, we summarize these results.

\paragraph{Error compensation methods outperform channel-wise transformations in most scenarios.} We observe that error compensation methods~(\textit{e.g.}, GPTQ and MR-GPTQ) consistently outperform channel-wise transformation methods~(\textit{e.g.}, SmoothQuant and AWQ), except in the case of Llama-3.1-8B-Instruct under W4A4. For example, based on Table~\ref{tab:pangu_nonreasoning}, under W4A8, error compensation methods recover 97.03\%–97.36\% of BF16 performance, outperforming channel-wise transformations (96.25\%–96.33\%). This performance gap may stem from the following: channel-wise scaling operates at a coarser granularity and cannot fully capture intra-group magnitude variations under MXFP’s group-wise quantization. Although error compensation methods like GPTQ also operate primarily at the channel level, they explicitly minimize quantization error during calibration, thereby providing a stronger guarantee.

\paragraph{Rotational transformation impairs MXFP4 quantization.} Unlike remarkable success in INT4 formats, Rotation-based methods (\textit{e.g.}, QuaRot, SpinQuant) consistently degrade MXFP4 quantization accuracy, performing worse than the RTN baseline. Specifically, Table~\ref{tab:llama_nonreasoning} shows that RTN achieves a lower perplexity (8.27) compared to QuaRot (10.34) and SpinQuant (10.40) under W4A4. On non-reasoning benchmarks, QuaRot and SpinQuant exhibit relative recovery rate degradations of 6.20\% and 7.73\% compared to RTN, respectively. The harm likely stems from the fact that MXFP4 uses group-wise scaling and relies on local statistical properties (e.g., distribution shape within each group) to preserve information during quantization. Global rotations mix information across all dimensions, flattening outlier structures and reducing kurtosis, which makes the distribution less amenable to effective group-wise scaling.

\paragraph{Affine transformation is most robust under 4-bit quantization.} Across all LLMs, affine transformation~(\textit{i.e.}, FlatQuant) exhibits the most robust performance under W4A4. For example, on Llama-3.1-8B-Instruct~(see Table~\ref{tab:llama_nonreasoning}), FlatQuant attains an accuracy recovery rate of 96.57\%, clearly surpassing RTN (94.98\%), SpinQuant (87.25\%), and even the strong error compensation method MR-GPTQ (95.65\%), while also achieving the lowest WikiText perplexity (8.03). The difference likely arises from the fact that, whereas orthogonal rotations preserve the $L_2$ norm, FlatQuant employs learnable affine transformations that do not conserve total energy. Although the global affine transforms of FlatQuant may still propagate energy across groups, their ability to modulate absolute magnitudes makes them better suited for low-bit MXFP quantization. To better understand how these methods reshape activation distributions, we visualize activation distributions under W4A4 in Figure~\ref{fig:activation_vis_layer8_qproj}. As illustrated in the figure, different quantization strategies induce markedly distinct activation distributions. Rotation-based approaches still preserve a relatively large activation magnitude compared to FlatQuant.

\paragraph{RTN remains a strong baseline across all bit-widths.} The experimental phenomenon underscores a critical limitation: most existing PTQ methods were designed and optimized for the INT format. However, these methods frequently fail to align with the quantization scheme of the MXFP format, yielding only marginal gains, or even regressions, over RTN when naively transferred to MXFP. This suggests that effective quantization methods for the MXFP format require designs that explicitly account for its quantization scheme.

\begin{mybox}[
  colback=gray!10,
  title=\textbf{Takeaways~2},
  boxrule=0.8pt,
  arc=2mm
]

Error compensation and affine transformations are better aligned with MXFP quantization, especially at low bit-widths. The persistence of RTN as a strong baseline suggests that MXFP requires quantization methods designed for its specific scheme.
\end{mybox}

\subsection{Impact of Model Families and Modalities~(RQ3)}
\label{sec:rq3}
To assess how quantization performance varies across model families and modalities, we evaluate models along two dimensions: (i) models with different modalities (\textit{i.e.}, Llama-3.1-8B-Instruct vs. Qwen2.5-VL-7B), and (ii) models of the same modality but from different families (\textit{i.e.}, Llama-3.1-8B-Instruct vs. openPangu-Embedded-7B-V1.1).

\begin{figure*}[!t]
  \centering
  \begin{minipage}[t]{0.58\textwidth}
    \centering
    \includegraphics[width=\linewidth]{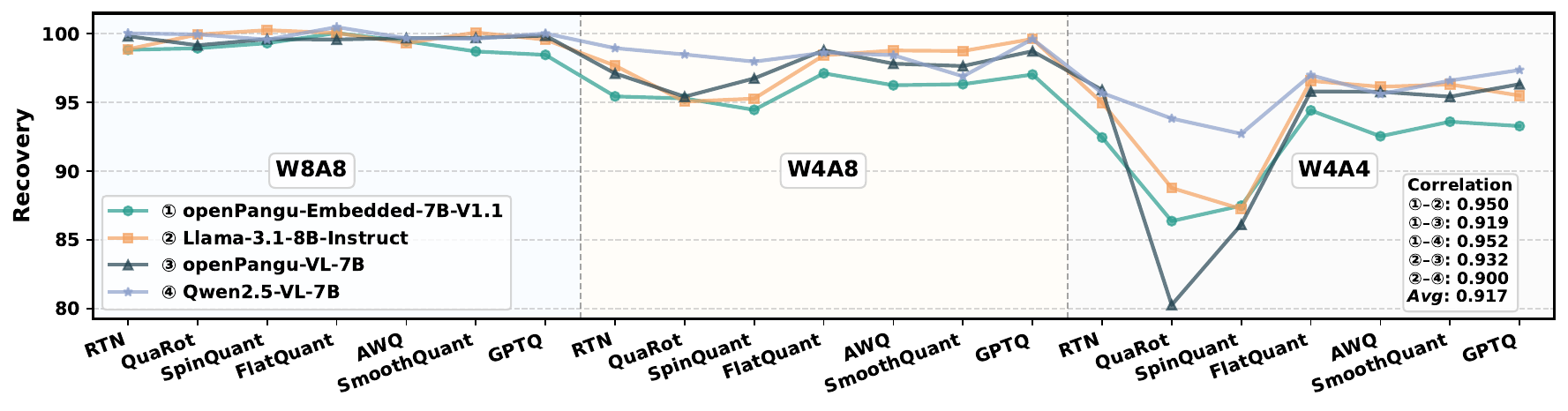}
    \caption{Recovery rate performance of various PTQ methods across MXFP quantization settings. Each curve represents results from all quantization settings of a model. We also show the pairwise correlation of two different models. The correlation is calculated using the Pearson correlation coefficient.}
    \label{fig:ptq_relative}
  \end{minipage}
  \hfill
  \begin{minipage}[t]{0.40\textwidth}
    \centering
    \includegraphics[width=\linewidth]{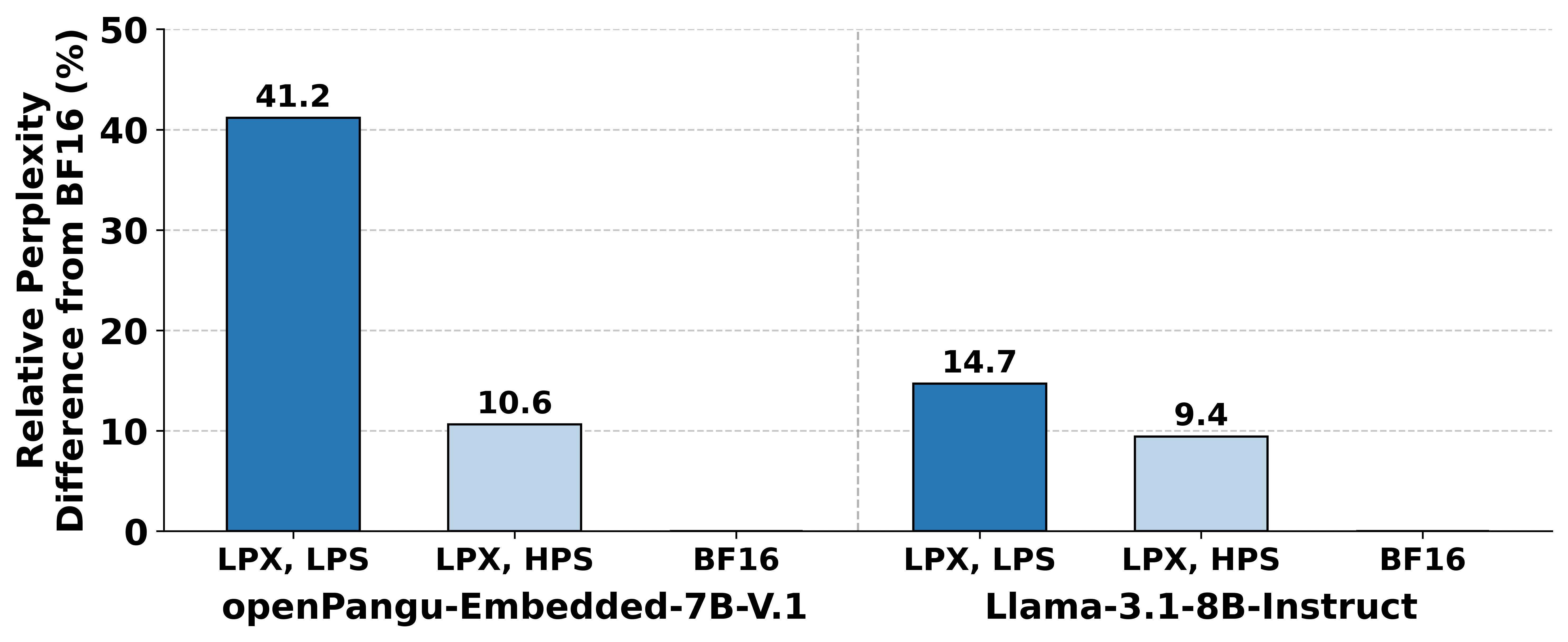}
    \caption{Impact of scaling factor error. Restoring low-precision values and low-precision scales (LPX, LPS) to low-precision values with high-precision scales (LPX, HPS) leads to a clear reduction in perplexity (PPL).}
    \label{fig:scaling_factor_error}
  \end{minipage}
  \vspace{-0.05cm}
\end{figure*}

\paragraph{PTQ effectiveness under MXFP is largely consistent across models and modalities.} Figure~\ref{fig:ptq_relative} summarizes the recovery rate performance of four models under different MXFP quantization settings and PTQ methods. We further analyze the correlation (the Pearson correlation coefficient~\citep{benesty2009pearson}) among their performance curves and observe a high degree of consistency, with an average pairwise correlation of 0.917. This strong alignment indicates that PTQ methods under MXFP exhibit similar performance trends across architectures and modalities, suggesting that their effectiveness is not strongly model- or modality-dependent.

\input{tables/vl_llm_vit}

\paragraph{In MLLMs, quantization sensitivity is dominated by the LLM rather than vision Transformer.} To dissect the contribution of each component of MLLMs to final results, we separately quantize the LLM and vision Transformer~(ViT) in Qwen2.5-VL-7B using RTN. As seen in Table~\ref{tab:vl_quant_comparison}, reducing the LLM from W8A8 to W4A4 induces a significant accuracy recovery rate drop (3\% in Qwen2.5-VL-7B), whereas applying the same W4A4 quantization to the ViT results in approximately 1\% degradation.
These results suggest a practical and effective quantization policy: retain higher precision (\textit{e.g.}, W8A8) in the LLM while aggressively quantizing the ViT (\textit{e.g.}, W4A4). Under this configuration, the model achieves nearly lossless accuracy while reducing memory footprint and the latency of the first token generation, making it a compelling default for realistic multimodal deployments.

\input{tables/vl_textual_visual}
\paragraph{Visual tokens are more robust under MXFP quantization compared to INT formats.}
Prior studies have shown that quantizing MLLMs to INT formats is particularly challenging, as visual tokens exhibit significantly larger outliers and a wider activation range than their textual counterparts~\citep{yu2025mquantunleashinginferencepotential,xue2025vlmqefficientposttrainingquantization}. However, as shown in Table~\ref{tab:vl_textual_visual}, under MXFP quantization, reducing the bit-width allocated to visual tokens does not result in noticeable accuracy degradation. This observation may be associated with MXFP’s exponent–mantissa decoupling, which allows more flexible handling of wide activation ranges while maintaining enough precision.

\begin{mybox}[
  colback=gray!10,
  title=\textbf{Takeaways~3},
  boxrule=0.8pt,
  arc=2mm
]

PTQ methods under MXFP exhibit consistent effectiveness across models and modalities. In MLLMs, quantization sensitivity is dominated by the LLM rather than the ViT, favoring a mixed-precision design that preserves higher LLM precision. Visual tokens are more robust under MXFP than INT, with reduced bit-width causing no noticeable accuracy loss.
\end{mybox}

\subsection{Analysis of MXFP4 Quantization Components~(RQ4)}
\label{sec:rq4}
\input{tables/pre_scale}

\paragraph{Quantization Error From Scaling Factors.} Considering the non-trivial quantization error introduced by MXFP4 quantization, we study the error introduced by the scaling factors. FP8 block-level scaling factors $s$ in MXFP formats must satisfy the constraints of the E8M0 data type. Following~\citep{cook2025four}, we keep the scaling factors in high precision while still mapping the scaled values onto the FP4 representable grid, and compute the relative perplexity (PPL) difference from BF16, as shown in Figure~\ref{fig:scaling_factor_error}. We observe that errors from scaling factors have a significant impact on model performance. When the error of scaling factors is addressed, the relative perplexity (PPL) improves clearly. This impact stems from the E8M0 format, which forces scaling factors to be powers of two. This coarse quantization often results in a large mismatch between the optimal scale and the allowed scale, thereby affecting all values in its block.

\paragraph{Pre-Scale Strategy.} 
As pointed out in~\citep{tseng2025training}, we adopt the unbiased MXFP4 quantization strategy to mitigate systematic clipping bias introduced by the limited FP4 dynamic range. Specifically, the input is first scaled by a factor of 3/4 before quantization, effectively preventing clipping while preserving relative magnitudes. As shown in Table~\ref{tab:pre_scale}, enabling \textit{Pre-scale} operation enhances performance from 52.39\% to 56.76\% and reduces PPL from 104.42 to 49.33.

\begin{mybox}[
  colback=gray!10,
  title=\textbf{Takeaways~4},
  boxrule=0.8pt,
  arc=2mm
]

The quantization error from scaling factors is not negligible. The \textit{Pre-scale} optimization strategy is recommended.
\end{mybox}

%% file: tables/llama8b.tex

\begin{table*}[!t]
\centering
\resizebox{0.75\linewidth}{!}{
\begin{tabular}{llccccccc|c}
\toprule[1.5pt]
\multirow{2}{*}{\textbf{Bits}} &
\multirow{2}{*}{\textbf{Method}} &
\multicolumn{7}{c|}{\textbf{ACC (0-shot) ↑}} &
\textbf{PPL ↓} \\
\cmidrule(lr){3-9} \cmidrule(l){10-10}
&
& \textbf{ARC-C} & \textbf{ARC-E} & \textbf{HellaSwag} & \textbf{PIQA} & \textbf{Winogrande} & \textit{\textbf{Avg.}} & \textit{\textbf{Recovery (\%)}} & \textbf{WikiText} \\
\midrule

BF16 & -- & 55.20 & 79.63 & 79.15 & 81.07 & 73.95 & 73.80 & 100.00 & 7.21 \\

\cmidrule{1-10}

\multirow{11}{*}{W8A8}
 & RTN               & 53.50 & 78.45 & 79.11 & 80.14 & 73.64 & 72.97 & \cellcolor{medium}98.87 & 7.31 \\
 & QuaRot            & 55.72 & 80.09 & 78.76 & 80.58 & 73.64 & 73.76 & \cellcolor{best}99.94 & 7.39 \\
 & $\text{QuaRot}^{*}$            & 56.4 & 80.81 & 78.84 & 81.01 & 74.35 & \textbf{74.28} & \cellcolor{best}\textbf{100.65} & 7.40 \\
 & SpinQuant         & 55.89 & 80.18 & 78.60 & 80.79 & 74.51 & 73.99 & \cellcolor{best}100.26 & 7.39 \\
 & $\text{SpinQuant}^{*}$            & 54.35 & 80.09 & 78.12 & 80.2 & 73.8 & 73.31 & \cellcolor{best}99.34 & 7.41 \\
 & FlatQuant         & 55.80 & 79.46 & 78.99 & 81.12 & 73.80 & 73.83 & \cellcolor{best}100.05 & 7.27 \\
 & $\text{FlatQuant}^{*}$            & 54.69 & 79.42 & 78.56 & 80.63 & 74.19 & 73.50 & \cellcolor{best}99.59 & 7.33 \\
 & AWQ               & 54.27 & 79.17 & 78.48 & 80.74 & 73.80 & 73.29 & \cellcolor{best}99.31 & 7.34 \\
 & SmoothQuant       & 55.38 & 79.08 & 79.14 & 81.23 & 74.51 & 73.87 & \cellcolor{best}100.09 & 7.34 \\
 & $\text{SmoothQuant}^{*}$            & 55.03 & 79.63 & 78.73 & 81.39 & 73.88 & 73.73 & \cellcolor{best}99.91 & 7.34 \\
 & MR-GPTQ           & 55.38 & 79.12 & 78.99 & 81.01 & 73.95 & 73.69 & \cellcolor{best}99.85 & \textbf{7.24} \\
 & GPTQ              & 54.69 & 79.12 & 78.47 & 80.79 & 74.35 & 73.48 & \cellcolor{best}99.57 & 7.33 \\

\cmidrule{1-10}

\multirow{11}{*}{W4A8}
 & RTN               & 53.07 & 76.81 & 77.04 & 80.03 & 73.48 & 72.09 & \cellcolor{medium}97.68 & \textbf{7.71} \\
 & QuaRot            & 49.15 & 75.93 & 76.37 & 78.29 & 71.03 & 70.15 & \cellcolor{worst}95.06 & 8.45 \\
 & $\text{QuaRot}^{*}$            & 52.65 & 79.67 & 77.86 & 80.52 & 73.40 & 72.82 & \cellcolor{medium}98.67 & 7.95 \\
 & SpinQuant         & 49.91 & 75.59 & 76.10 & 78.07 & 71.90 & 70.31 & \cellcolor{worst}95.28 & 8.50 \\
 & $\text{SpinQuant}^{*}$            & 52.82 & 78.37 & 76.96 & 79.6 & 73.32 & 72.21& \cellcolor{medium}97.85 & 7.84 \\
 & FlatQuant         & 53.84 & 80.26 & 77.27 & 80.03 & 71.82 & 72.64 & \cellcolor{medium}98.43 & 7.87 \\
 & $\text{FlatQuant}^{*}$            & 53.84 & 80.26 & 77.27 & 80.03 & 71.82 & 72.64 & \cellcolor{medium}97.95 & 7.81 \\
 & AWQ               & 53.84 & 79.25 & 77.55 & 79.98 & 73.88 & 72.90  & \cellcolor{medium}98.78 & 7.75 \\
 & SmoothQuant       & 54.01 & 77.61 & 77.77 & 79.98 & 74.98 & 72.87  & \cellcolor{medium}98.74 & 7.75 \\
 & $\text{SmoothQuant}^{*}$            & 52.96 & 77.02 & 77.42 & 78.63 & 70.64 & 71.33 & \cellcolor{worst}96.67 &  8.70\\
 & MR-GPTQ           & 54.52    & 78.62    & 77.28    & 80.47    & 73.80    & 72.94    & \cellcolor{medium}98.83 & 7.83 \\
 & GPTQ              & 53.5 & 79.59 & 79.71 & 81.18 & 73.6 & \textbf{73.52} & \cellcolor{best}\textbf{99.62} & 7.87 \\

\cmidrule{1-10}

\multirow{11}{*}{W4A4}
 & RTN               & 49.66 & 75.80 & 75.48 & 79.05 & 70.48 & 70.09 & \cellcolor{worst}94.98 & 8.27 \\
 & QuaRot            & 44.11 & 71.63 & 71.82 & 75.14 & 64.88 & 65.52 & \cellcolor{worst}88.78 & 10.34 \\
 & $\text{QuaRot}^{*}$            & 49.15 & 74.62 & 74.26 & 77.48 & 68.98 & 68.90 & \cellcolor{worst}93.36 & 9.12 \\
 & SpinQuant         & 43.09 & 67.26 & 70.71 & 74.92 & 65.98 & 64.39 & \cellcolor{worst}87.25 & 10.40 \\
 & $\text{SpinQuant}^{*}$            & 49.15 & 74.54 & 74.38 & 76.99 & 69.53 & 68.92 & \cellcolor{worst}93.38 & 9.01 \\
 & FlatQuant         & 52.05 & 77.27 & 76.89 & 79.49 & 70.64 & 71.27 & \cellcolor{worst}96.57 & \textbf{8.03} \\
 & $\text{FlatQuant}^{*}$            & 51.19 & 78.58 & 76.77 & 78.94 & 71.67 & \textbf{71.43} & \cellcolor{worst}\textbf{96.79} & 8.06 \\
 & AWQ               & 52.30 & 77.95 & 76.13 & 78.92 & 69.46 & 70.95  & \cellcolor{worst}96.14 & 8.25 \\
 & SmoothQuant       & 51.37 & 76.52 & 76.12 & 79.00 & 72.38 & 71.08  & \cellcolor{worst}96.31 & 8.25 \\
 & $\text{SmoothQuant}^{*}$            & 50.51 & 73.36 & 75.27 & 76.82 & 67.17 & 68.63 & \cellcolor{worst}92.98 & 8.50 \\
 & MR-GPTQ           & 50.85 & 75.46 & 76.02 & 79.82 & 70.80 & 70.59 & \cellcolor{worst}95.65 & 8.34 \\
 & GPTQ              & 50.68 & 77.65 & 75.66 & 78.18 & 70.24 & 70.48 & \cellcolor{worst}95.50 & 8.37 \\

\bottomrule[1.5pt]
\end{tabular}
}
\vspace{-0.25cm}
\caption{Comparison of PTQ methods on \textbf{Llama-3.1-8B-Instruct} under W8A8, W4A8, and W4A4 in terms of \textbf{non-reasoning} downstream task accuracy and perplexity. * denotes the variant integrated with the GPTQ algorithm.}
\vspace{-0.25cm}
\label{tab:llama_nonreasoning}
\end{table*}

%% file: tables/pangu7b.tex
\begin{table*}[!t]
\centering
\resizebox{0.75\linewidth}{!}{

\begin{tabular}{llccccccc|c}
\toprule[1.5pt]
\multirow{2}{*}{\textbf{Bits}} &
\multirow{2}{*}{\textbf{Method}} &
\multicolumn{7}{c|}{\textbf{ACC (0-shot) ↑}} &
\textbf{PPL ↓} \\ \cmidrule(lr){3-9} \cmidrule(l){10-10}
&
&
\multicolumn{1}{l}{\textbf{ARC-C}} &
\multicolumn{1}{l}{\textbf{ARC-E}} &
\multicolumn{1}{l}{\textbf{HellaSwag}} &
\multicolumn{1}{l}{\textbf{PIQA}} &
\multicolumn{1}{l}{\textbf{Winogrande}} &
\textit{\textbf{Avg.}} &
\textit{\textbf{Recovery.(\%)}} &
\multicolumn{1}{l}{\textbf{WikiText}} \\ \midrule

BF16                      & --               & 42.75 & 67.26 & 62.86 & 73.50 & 60.62 & 61.40 & 100.00 & 34.89 \\ \cmidrule{1-10} 
\multirow{11}{*}{W8A8}    & RTN              & 44.03 & 66.25 & 61.74 & 73.61 & 57.77 & 60.68 & \cellcolor{medium}{98.83} & 35.75 \\
                          & QuaRot           & 42.24 & 65.95 & 62.45 & 72.42 & 60.69 & 60.75 & \cellcolor{medium}98.94 & 35.21 \\
                          & $\text{QuaRot}^{*}$           & 42.83 & 68.22 & 62.18 & 73.78 & 60.96 & 61.59 & \cellcolor{best}{100.32} & 35.75 \\
                          & SpinQuant        & 42.49 & 67.34 & 62.42 & 72.31 & 60.30 & 60.97 & \cellcolor{best}{99.31} & 35.49 \\
                          & $\text{SpinQuant}^{*}$  & 42.58 & 68.43 & 62.52 & 73.94 & 59.59 & 61.41 & \cellcolor{best}{100.02} & 33.04 \\
                          & FlatQuant        & 43.43 & 68.64 & 62.25 & 73.78 & 59.04 & 61.43 & \cellcolor{best}{100.05} & {30.96} \\
                          & $\text{FlatQuant}^{*}$ & 43.86 & 69.23 & 62.33 & 74.05 & 61.33 & \textbf{62.16} & \cellcolor{best}{\textbf{101.24}} & \textbf{30.87} \\
                          & AWQ              & 41.98 & 68.48 & 61.16 & 72.58 & 61.09 & 61.06 & \cellcolor{best}{99.45} & 38.00 \\
                          & SmoothQuant      & 42.58 & 66.50 & 62.08 & 72.69 & 59.19 & 60.61 & \cellcolor{medium}98.71 & 35.25 \\
                          & $\text{SmoothQuant}^{*}$ & 42.15 & 66.67 & 61.84 & 72.36 & 59.59 & 60.52 & \cellcolor{medium}{98.57} & {35.00} \\
                          & MR-GPTQ          & 43.09 & 67.80 & 62.84 & 73.72 & 59.91 & {61.47} & \cellcolor{best}{100.12}  & 34.57                     \\
                          & GPTQ             & 42.92 & 67.13 & 61.66 & 72.69 & 57.85 & 60.45 & \cellcolor{medium}98.46 & 34.75 \\ \cmidrule{1-10} 

\multirow{11}{*}{W4A8}    & RTN              & 39.59 & 65.24 & 58.70 & 71.60 & 57.85 & 58.60 & \cellcolor{worst}95.44 & 42.17 \\
                          & QuaRot           & 40.10 & 64.10 & 57.95 & 72.69 & 57.70 & 58.51 & \cellcolor{worst}95.29 & 40.87 \\
                          & $\text{QuaRot}^{*}$ & 40.02 & 65.07 & 58.54 & 71.44 & 57.62 & 58.54 & \cellcolor{worst}95.84 & 43.83 \\
                          & SpinQuant        & 39.59 & 61.57 & 58.56 & 72.09 & 58.17 & 58.00 & \cellcolor{worst}94.46 & 38.94 \\
                          & $\text{SpinQuant}^{*}$ & 42.41 & 67.85 & 60.74 & 72.91 & 58.8 & \textbf{60.54} & \cellcolor{medium}\textbf{98.60} & \textbf{34.75} \\
                          & FlatQuant        & 40.78 & 66.54 & 60.15 & 72.52 & 58.17 & 59.63 & \cellcolor{medium}97.12 & 37.67 \\
                          & $\text{FlatQuant}^{*}$ & 42.15 & 67.63 & 60.54 & 72.80 & 57.38 & 60.10 & \cellcolor{medium}97.89 & 37.72 \\
                          & AWQ              & 39.85 & 66.25 & 58.13 & 73.07 & 58.17 & 59.09 & \cellcolor{worst}96.25 & 41.75 \\
                          & SmoothQuant      & 40.78 & 66.20 & 58.82 & 72.47 & 57.46 & 59.15 & \cellcolor{worst}96.33 & 43.25 \\
                          & $\text{SmoothQuant}^{*}$ & 41.64 & 67.47 & 59.88 & 73.07 & 58.88 & 60.19 & \cellcolor{medium}98.03 & 40.00 \\

                          & MR-GPTQ          & 40.87 & 66.84 & 60.64 & 72.63 & 57.93 & {59.78} & \cellcolor{medium}{97.36} &       39.19                 \\
                          & GPTQ             & 40.96 & 65.51 & 59.48 & 72.42 & 59.51 & 59.58 &  \cellcolor{medium}97.03 & {37.50}  \\ \cmidrule{1-10} 

\multirow{11}{*}{W4A4}    & RTN              & 38.48 & 61.53 & 56.42 & 70.46 & 56.91 & 56.76 & \cellcolor{worst}92.45 & 49.33 \\
                          & QuaRot           & 36.43 & 56.19 & 51.24 & 66.76 & 54.54 & 53.03 & \cellcolor{worst}86.37 & 56.19 \\
                          & $\text{QuaRot}^{*}$ & 36.09 & 59.30 & 53.06 & 68.55 & 55.33 & 54.47& \cellcolor{worst}88.71 & 53.75 \\
                          & SpinQuant        & 34.64 & 57.24 & 52.82 & 68.39 & 55.49 & 53.72 & \cellcolor{worst}87.49 & 51.09 \\
                          & $\text{SpinQuant}^{*}$ & 37.88 & 60.4 & 55.34 & 69.53 & 57.22 & 56.07 & \cellcolor{worst}91.33 & 46.28 \\
                          & FlatQuant        & 39.93 & 65.82 & 57.07 & 71.00 & 56.04 & {57.97} & \cellcolor{worst}{94.42} & {38.36} \\
                          & $\text{FlatQuant}^{*}$ & 40.19 & 66.84 & 58.13 & 69.80 & 57.54 & \textbf{58.50} & \cellcolor{worst}\textbf{95.28} & \textbf{36.40} \\
                          & AWQ              & 37.08 & 63.76 & 54.49 & 71.06 & 57.70 & 56.82 & \cellcolor{worst}92.54 & 46.00 \\
                          & SmoothQuant      & 38.99 & 65.07 & 55.81 & 70.80 & 56.70 & 57.47 & \cellcolor{worst}93.60 & 52.00 \\
                          & $\text{SmoothQuant}^{*}$ & 38.65 & 64.60 & 56.74 & 70.73 & 57.62 & 57.67 & \cellcolor{worst}93.92 & 43.57 \\
                          & MR-GPTQ          & 38.99 & 63.43 & 57.19 & 69.48 & 58.96 & 57.61 & \cellcolor{worst}93.83 & 42.17                       \\
                          & GPTQ             & 39.25 & 62.75 & 57.47 & 69.91 & 56.99 & 57.27 & \cellcolor{worst}93.28 & 44.50 \\ \bottomrule[1.5pt]
\end{tabular}
}
\caption{Comparison of PTQ methods on \textbf{openPangu-Embedded-7B-V1.1} under W8A8, W4A8, and W4A4  about \textbf{non-reasoning} downstream task accuracy and perplexity. * denotes the variant integrated with the GPTQ algorithm.}
\label{tab:pangu_nonreasoning}
\end{table*}

%% file: tables/qwen_vl.tex
\begin{table*}[!t]
\renewcommand{\arraystretch}{1.0}
\centering
\resizebox{0.8\linewidth}{!}{
\begin{tabular}{llccccccc|c}
\toprule[1.5pt]
\textbf{Bits} & \textbf{Method} &
\textbf{OCRBench} &
\textbf{MMBench} &
\textbf{$\text{MMBench}^{CN}$} &
\textbf{TextVQA} &
\textbf{ChartQA} &
\textbf{MME} &
\textbf{MMMU} &
\textit{\textbf{Recovery~(\%)}}  \\
\cmidrule(lr){1-9} \cmidrule(l){10-10}
BF16 & --               & 877 &	79.08 &	75.95 &	84.61 &	85.61 &	2321 &	49.5 &	100.00%
 \\ 
\cmidrule{1-10} 

\multirow{7}{*}{W8A8} & RTN           & 879 & 78.32 & 75.34	 & 84.32 &	86.68 &	2312 & 50.1 & \cellcolor{best}100.03  \\
     & QuaRot        & 874 &	79.00 &	75.86 &	84.21 &	86.20 &	2302 &	49.9 & \cellcolor{best}99.95 \\
     & QuaRot*       & 873 &	78.32 &	74.48 &	83.90 &	86.60 &	2300 &	49.2 & \cellcolor{best}99.35 \\
     & SpinQuant     & 880 & 78.66 & 74.74 & 84.33 & 86.32 & 2323 & 48.6 & \cellcolor{best}99.57 \\
     & SpinQuant*    & 881 & 79.12 & 75.16 & 84.63 & 86.42 & 2325 & 49.4 & \cellcolor{best} 100.06 \\
     & FlatQuant     & 879 &	79.17 &	75.95 &	84.63 &	86.64 &	2329 &	50.2 & \cellcolor{best}\textbf{100.48} \\
     & FlatQuant*    & 878 & 78.66 & 75.69 & 84.51 & 86.28 & 2339 & 50.3 & \cellcolor{best} 100.33 \\
     & AWQ           & 873 &	78.15 &	76.03 &	83.46 &	86.16 &	2300 & 49.9
& \cellcolor{best}99.67 \\
     & SmoothQuant   & 874 &	78.32 &	75.09 &	83.73 &	86.28 &	2337 &	49.2 & \cellcolor{best}99.63 \\
     & SmoothQuant*  & 883 &	78.49 &	76.37 &	83.75 &	85.40 &	2314 &	50.3 & \cellcolor{best}100.08 \\
     & GPTQ          & 872 &	78.49 &	76.29 &	84.29 &	85.81 &	2315 &	50.2 & \cellcolor{best}{100.02} \\ 
\cmidrule{1-10} 

\multirow{7}{*}{W4A8} & RTN           & 855 & 78.57 & 76.80 & 82.74 &	84.56 & 2240 & 50.3 & \cellcolor{medium}98.95 \\
     & QuaRot        & 863 &	77.98 &	73.63 &	82.87 &	85.08 &	2297 &	49.1 & \cellcolor{medium}98.49 \\
     & QuaRot*       & 859 &	79.00 &	76.55 &	82.91 &	84.60 &	2208 &	49.2 & \cellcolor{medium}98.57 \\
     & SpinQuant     & 864 &	77.89 &	74.40 &	83.30 &	85.68 &	2137 &	49.6 & \cellcolor{medium}97.97 \\
     & SpinQuant*    & 866 &	78.40 &	73.97 &	83.81 &	85.72 &	2271 &	50.9 & \cellcolor{best}99.31 \\
     & FlatQuant     & 861 &	76.70 &	75.77 &	81.81 &	86.28 &	2287 &	49.2 & \cellcolor{medium}98.62 \\
     & FlatQuant*    & 870 &	76.87 &	73.59 &	82.96 &	86.28 &	2307 &	49.5 & \cellcolor{medium}98.79 \\
     & AWQ           & 866 &	77.72 &	74.31 &	81.82 &	85.48 &	2314 &	48.5 & \cellcolor{medium}98.44 \\
     & SmoothQuant   & 849 &	76.19 &	74.44 &	81.1 &	84.36 &	2242 &	47.6 & \cellcolor{worst}96.90 \\
     & SmoothQuant*  & 868 &	78.32 &	75.43 &	83.45 &	85.08 &	2275 &	48.7 & \cellcolor{medium}98.82 \\
     & GPTQ          & 864 &	77.89 &	76.03 &	83.18 &	85.80 &	2325 &	50.2 & \cellcolor{best}\textbf{99.61} \\ 
\cmidrule{1-10} 

\multirow{7}{*}{W4A4} & RTN           & 845 & 77.04 & 74.88 & 78.62	& 82.97 &	2194 & 46.1 & \cellcolor{worst}95.69 \\
     & QuaRot        & 843 &	73.83 &	68.73 &	78.89 &	81.76 &	2195 &	46.3 & \cellcolor{worst}93.83 \\
     & QuaRot*       & 840 &	75.34 &	71.91 &	80.20 &	84.56 &	2103 &	47.0 & \cellcolor{worst}94.98 \\
     & SpinQuant     & 843 &	73.21 &	69.17 &	78.98 &	80.76 &	2006 &	47.1 & \cellcolor{worst}92.72 \\
     & SpinQuant*    & 851 &	74.83 &	70.19 &	80.29 &	82.60 &	2190 & 47.2 & \cellcolor{worst}95.02 \\
     & FlatQuant     & 853 &	75.26 &	74.31 &	80.14 &	85.44 &	2293 &	47.2 & \cellcolor{worst}96.99  \\
     & FlatQuant*    & 860 &	74.23 &	69.67 &	81.56 &	86.32 &	2287 
     & 48.4 & \cellcolor{worst}96.74  \\
     & AWQ           & 840 &	74.83 &	72.08 &	79.71 &	83.32 &	2249 &	47.3 & \cellcolor{worst}95.61 \\
     & SmoothQuant   & 843 &	75.26 &	72.51 &	81.26 &	84.04 &	2204 &	49.6 & \cellcolor{worst}96.59  \\
     & SmoothQuant*  & 843 &	76.62 &	73.20 &	81.35 &	84.60 &	2231 &	47.0 & \cellcolor{worst}96.49  \\
     & GPTQ          & 846 &	76.19 &	73.97 &	81.12 &	83.72 &	2336 &	48.0 & \cellcolor{medium}\textbf{97.36}  \\ 
\bottomrule[1.5pt]
\end{tabular}
}
\vspace{-0.25cm}
\caption{Comparison of PTQ methods on \textbf{Qwen2.5-VL-7B} under W8A8, W4A8, and W4A4 quantization across multimodal benchmarks. * denotes the variant integrated with the GPTQ algorithm. 
}
\vspace{-0.25cm}
\label{tab:qwenvl}
\end{table*}

%% file: tables/vl_llm_vit.tex
\begin{table}[!t]
\centering

\resizebox{\linewidth}{!}{
\begin{tabular}{l l c c c c c c c | c}
\toprule
\textbf{LLM} & \textbf{ViT} &
\textbf{OCRBench} & \textbf{MMBench} & \textbf{$\text{MMBench}^{CN}$} & \textbf{TextVQA} & \textbf{ChartQA} &
\textbf{MME} & \textbf{MMMU} & \textit{\textbf{Recovery~(\%)}} \\
\midrule
 BF16 & BF16 & 877 & 79.08 & 75.95 & 84.61 & 85.61 & 2321 & 49.5 & {100.00} \\
\midrule
\multirow{2}{*}{W8A8} & \cellcolor{int-bg} W4A4 & \cellcolor{int-bg} 856 & \cellcolor{int-bg} 77.72 & \cellcolor{int-bg} 75.43 & \cellcolor{int-bg} 83.13 & \cellcolor{int-bg} 85.56 & \cellcolor{int-bg} 2288 & \cellcolor{int-bg} 49.3 & \cellcolor{int-bg} 98.96 \\
 & \cellcolor{mxfp-bg} W8A8 & \cellcolor{mxfp-bg} 879 & \cellcolor{mxfp-bg} 78.32 & \cellcolor{mxfp-bg} 75.34 & \cellcolor{mxfp-bg} 84.32 & \cellcolor{mxfp-bg} 86.68 & \cellcolor{mxfp-bg} 2312 & \cellcolor{mxfp-bg} 50.1 & \cellcolor{mxfp-bg} 100.03 \\
\midrule

 \multirow{2}{*}{W4A4} & \cellcolor{int-bg} W4A4 & \cellcolor{int-bg} 845 & \cellcolor{int-bg} 77.04 & \cellcolor{int-bg} 74.88 & \cellcolor{int-bg} 78.62 & \cellcolor{int-bg} 82.97 & \cellcolor{int-bg} 2194 & \cellcolor{int-bg} 46.1 & \cellcolor{int-bg} 95.69 \\
 & \cellcolor{mxfp-bg} W8A8 & \cellcolor{mxfp-bg} 854 & \cellcolor{mxfp-bg} 77.81 & \cellcolor{mxfp-bg} 76.62 & \cellcolor{mxfp-bg} 80.50 & \cellcolor{mxfp-bg} 82.56 & \cellcolor{mxfp-bg} 2214 & \cellcolor{mxfp-bg} 47.9 & \cellcolor{mxfp-bg} 97.20 \\
\bottomrule
\end{tabular}
}
\vspace{-0.25cm}
\caption{
Comparison of the impact of the LLM and ViT quantization under RTN on Qwen-2.5-VL-7B. Recovery (\%) is computed relative to the BF16 baseline. 
}
\vspace{-0.25cm}
\label{tab:vl_quant_comparison}
\end{table}

%% file: tables/vl_textual_visual.tex
\begin{table}[!t]
\centering
\resizebox{\linewidth}{!}{
\begin{tabular}{l |l |l |c c c c c c c | c}
\toprule
\textbf{LLM Weights} & \textbf{Textual} & \textbf{Visual} &
\textbf{OCRBench} & \textbf{MMBench} & \textbf{$\text{MMBench}^{CN}$} & \textbf{TextVQA} & \textbf{ChartQA} &
\textbf{MME} & \textbf{MMMU} & \textit{\textbf{Recovery~(\%)}} \\
\midrule
\multirow{4}{*}{\textbf{W16}} 
 & A16 & A16 & 918 & 85.46 & 85.40 & 83.92 & 87.68 & 2287 & 54.5 & {100.00} \\
  & A16 & A4 & 901 & 85.37 & 85.14 & 83.68 & 86.96 & 2302 & 53.5 & 99.35 \\
   & A4 &A16 & 902 & 84.35 & 84.45 & 83.68 & 87.32 & 2335 & 53.4 & 99.32 \\
    & A4 &A4 & 908 & 83.59 & 82.13 & 83.42 & 87.36 & 2289 & 51.5 & 98.07 \\
\cmidrule{1-11}
\multirow{4}{*}{\textbf{W4}} 
 & A8 & A8 & 885 & 83.42 & 83.16 & 82.56 & 86.44 & 2270 & 53.2 & 97.89 \\
  & A8 & A4 &  878 & 82.91 & 83.16 & 82.26 & 86.52 & 2223 & 51.6 & 96.95 \\
   & A4 & A8 & 873 & 82.48 & 82.39 & 81.87 & 86.04 & 2268 & 51.4 & 96.75 \\
   & A4 & A4 & 873 & 82.48 & 82.56 & 81.39 & 86.00 & 2259 & 50.7 & 96.45 \\
\cmidrule{1-11}
\multirow{4}{*}{\textbf{W4*}} 
 & A8 & A8 & 899 & 85.29 & 83.85 & 83.4 & 87.28 & 2295 & 53.0 & 98.92 \\
  & A8 & A4 &  886 & 84.87 & 84.16 & 83.2 & 87.44 & 2316 & 53.6 & 98.98\\
   & A4 & A8 &  893 & 83.59 & 82.56 & 82.9 & 86.92 & 2288 & 51.9 & 97.85 \\
   & A4 & A4 &  876 & 82.91 & 81.96 & 83.06 & 86.88 & 2280 & 52.8 & 97.58 \\
\cmidrule{1-11}

\end{tabular}
}
\vspace{-0.05cm}
\caption{
Comparison of activation quantization for textual and visual tokens under different settings in openPangu-VL-7B. 
Weights are kept in full precision (W16) or quantized to 4-bit (W4) using RTN; the * indicates GPTQ-based weight quantization. 
Note that the ViT component remains unquantized.
}
\vspace{-0.15cm}
\label{tab:vl_textual_visual}
\end{table}

%% file: tables/pre_scale.tex
\begin{table}[!t]
\centering
\resizebox{\linewidth}{!}{
\begin{tabular}{l c ccccccc}
\toprule
& \multicolumn{1}{c}{\textbf{Strategy}} & \multicolumn{7}{c}{\textbf{Metric}} \\
\cmidrule(lr){2-2} \cmidrule(lr){3-9}
& \textit{Pre-scale} &  \textbf{ARC-C} & \textbf{ARC-E} & \textbf{HellaSwag} & \textbf{PIQA} & \textbf{Winogrande} & \textbf{Avg Acc $\uparrow$} & \textbf{PPL $\downarrow$} \\
\midrule
& \textcolor{origin}{\ding{51}} & 38.48 & 61.53 & 56.42 & 70.46 & 56.91 & 56.76 & 49.33 \\
& \textcolor{gray}{\ding{55}} & 32.34 &	58.12 & 48.76 & 67.41 & 55.33	& 52.39	& 104.42
\\

\bottomrule
\end{tabular}
}
\caption{Ablation study of optimization strategy for openPangu-Embedded-7B-V1.1 under W4A4. \textit{Pre-scale} refers to scaling inputs before computing quantization parameters.}
\label{tab:pre_scale}
\end{table}

%% file: latex/conclusion.tex
\section{Conclusion}
This work systematically studies PTQ under MXFP formats. We show that MXFP8 supports stable, near-lossless deployment. Quantization effectiveness under MXFP is governed by compatibility with its block floating-point scheme, resulting in consistent performance differences across algorithmic paradigms. Across model families and modalities, PTQ performance exhibits highly aligned trends. In multimodal models, quantization sensitivity is dominated by the language component. We further identify scaling-factor quantization as a key error source in MXFP4 and show that a simple pre-scale optimization strategy can effectively mitigate its impact. Overall, MXFP should be treated as a distinct numerical regime rather than a drop-in replacement for integer formats. Our findings provide practical guidance for MXFP-aware quantization design.

\section*{Limitation}
Our study is conducted on 7B/8B-scale LLMs and MLLMs and focuses on MXFP formats. While these settings cover widely used foundation models and representative microscaling designs, we do not evaluate substantially larger models~(\textit{e.g.}, 30B-scale models) or NVIDIA-specific NVFP formats~(\textit{e.g.}, NVFP4 and NVFP8). Although many observations appear to arise from intrinsic properties of block floating-point quantization and therefore suggest a degree of generality, it remains unclear how well these conclusions extend to much larger model scales or to alternative microscaling formats with different exponent and scaling designs. A systematic investigation along these directions is left for future work.

%% file: latex/appendix.tex
\section{Appendix}
\label{sec:appendix}
\subsection{Benchmarks and Evaluation Details}
\label{appendix:evaluation}
Below, we briefly introduce the benchmarks and the evaluation details in this study.

\paragraph{Non-Reasoning Benchmarks.} \textbf{PIQA}: It is a physical commonsense reasoning and corresponding benchmark dataset, which was designed to investigate the physical knowledge of existing models. \textbf{Winogrande}: Winogrande is a collection of 44k problems formulated as a fill-in-a-blank task with binary options, and the goal is to choose the right option for a given sentence, which requires commonsense reasoning. \textbf{Hellaswag}: It is a commonsense inference benchmark designed to challenge language
models with adversarially filtered multiple-choice questions. \textbf{ARC-Easy \& ARC-Challenge}: The ARC dataset consists of 7,787 science exam questions drawn from a variety of sources. Each question has a multiple-choice structure (typically 4 answer options). ARC-Easy contains 5,197 easy questions, and ARC-Challenge contains 2,590 hard questions. For all non-reasoning benchmarks, we use lm-evaluation-harness~\citep{eval-harness} with the vllm~\citep{kwon2023efficient} backend for evaluation. 

\paragraph{Reasoning Benchmarks.} \textbf{MATH500}: A benchmark that contains a mix of easy and hard mathematical problems designed to test comprehensive reasoning abilities. We evaluate model performance using \textbf{Avg@1} (\textit{i.e.}, the accuracy of the first generated answer). \textbf{AIME24}: It contains 30 problems from the American Invitational Mathematics Examination (AIME) 2024. \textbf{AIME25}: It contains 30 problems from the American Invitational Mathematics Examination (AIME) 2025. Following standard practice for high-stakes math benchmarks, we report results using \textbf{Avg@16} for AIME24 and AIME25, which averages accuracy over 16 independently sampled reasoning traces per problem. For all reasoning benchmarks, we follow~\citep{openPangu-Embedded-7B-v1.1} to evaluate openPangu-Embedded-7B-V1.1 without extra CoT prompts across all reasoning benchmarks. We use Lighteval~\citep{lighteval} with the vllm~\citep{kwon2023efficient} backend for evaluation with a sampling temperature of 1.0 and top-p of 0.8. The maximum sequence length of the model is limited to 131,072.

\paragraph{Image-Text Benchmarks.} \textbf{OCRBench}: OCRBench is a comprehensive evaluation benchmark designed to assess the OCR capabilities of Large Multimodal Models, which contains 1000 question-answer pairs, including Text Recognition, SceneText-Centric VQA, Document-Oriented VQA, Key Information Extraction, and Handwritten Mathematical Expression Recognition. \textbf{MMBench}: It is a collection of benchmarks to evaluate the multimodal understanding capability of large vision language models (LVLMs). \textbf{TextVQA}: TextVQA evaluates a model’s ability to read and reason about text present in images. We use the validation set, which contains 5,000 question-answer pairs. \textbf{ChartQA}: A benchmark focused on chart understanding and reasoning.  \textbf{MME}: A benchmark that evaluates LVLMs across multiple dimensions, including perception, cognition, and object hallucination.  \textbf{MMMU}: This is a challenging multidisciplinary problem involving benchmarking across fields such as art, engineering, law, and medicine. For all Image-Text benchmarks, we use metrics in LMMs-Eval~\citep{zhang2025lmms} with the vllm~\citep{kwon2023efficient} backend for evaluation. In the main tables, Vision Transformer (ViT) is quantized by RTN, and LLM is quantized by different PTQ methods.

\subsection{Additional Experiments}
\subsubsection{Results on Reasoning Benchmarks}
\label{appendix:reasoning}
\input{tables/pangu7b_reasoning}
We provide the detailed results on reasoning benchmarks in Table~\ref{pangu_math_reasoning}.

\subsubsection{Results on openPangu-VL-7B}
\label{appendix:pangu7bvl}
\input{tables/pangu_7bvl}
We provide the detailed evaluation results on openPangu-VL-7B for reference (see Table~\ref{tab:pangu7bvl}).

\subsubsection{Results under More Quantization Scenarios}
\label{appendix:more_senario}
\input{tables/pangu7b_all}
\input{tables/llama8b_all}
Below, we report the results of openPangu-Embedded-7B-V1.1 and Llama-3.1-8B-Instruct under more bit-width configurations (including W4A16 and W4A8KV8) as shown in Table \ref{pangu_nonreasoning_all} and Table \ref{llama_nonreasoning_all}, respectively.

\paragraph{Results on W4A16.} According to Table \ref{pangu_nonreasoning_all} and Table \ref{llama_nonreasoning_all}, even though activations remain at high precision (16-bit), quantizing weights to 4 bits already leads to a noticeable performance drop for most methods. For instance, QuaRot achieves only a 95.27\% recovery rate on openPangu-Embedded-7B-V1.1, falling into the \textcolor{red}{risky} regime ($<$97\%). Interestingly, when comparing W4A16 with the more aggressive W4A8 setting, we observe that the primary bottleneck lies in 4-bit weight quantization itself, rather than 8-bit activation compression, as the performance of most methods only slightly decreases. For example, on openPangu-Embedded-7B-V1.1, SpinQuant’s recovery rate drops only slightly from 95.10\% (W4A16) to 94.46\% (W4A8), while FlatQuant's recovery rate even improves from 96.87\%  (W4A16) to 97.12\% (W4A8). These findings suggest that even the 4-bit weight quantization alone constitutes a significant challenge.

\paragraph{Results on W4A8KV8.} We further investigated the impact of 8-bit KV cache quantization. According to Table \ref{pangu_nonreasoning_all} and Table \ref{llama_nonreasoning_all}, comparing the W4A8 setting with W4A8KV8, some methods maintain stable performance when the KV cache is 8-bit quantized. For example, on openPangu-Embedded-7B-V1.1, RTN, SpinQuant, and FlatQuant show only minor changes in accuracy recovery rate, changing from 95.44\%, 94.46\%, and 97.12\% to 95.14\%, 94.33\%, and 97.42\%, respectively. However, other methods exhibit noticeable degradation. For instance, the accuracy recovery rate of SmoothQuant and GPTQ drops from 96.33\% and 97.03\% to 94.96\% and 95.32\%, respectively. These results indicate that 8-bit KV cache quantization is not risk-free, and different quantization methods exhibit varying degrees of robustness to KV cache precision reduction.

\begin{figure*}[htbp]
\centering

\begin{subfigure}[b]{0.24\textwidth}
  \centering
  \includegraphics[width=\linewidth]{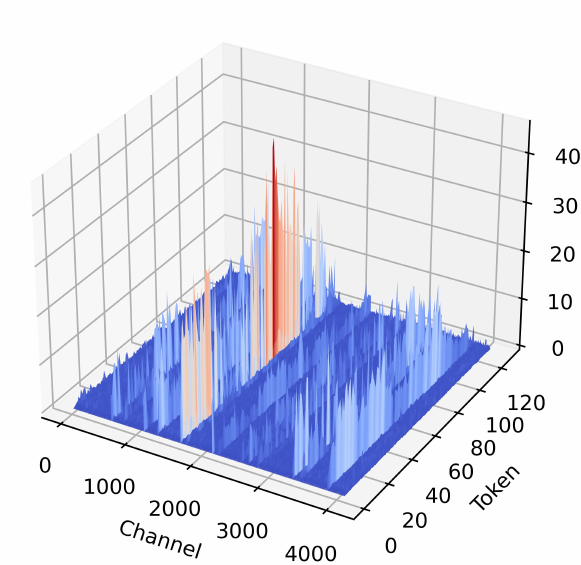}
  \caption{BF16}
  \label{fig:1}
\end{subfigure}
\hfill
\begin{subfigure}[b]{0.24\textwidth}
  \centering
  \includegraphics[width=\linewidth]{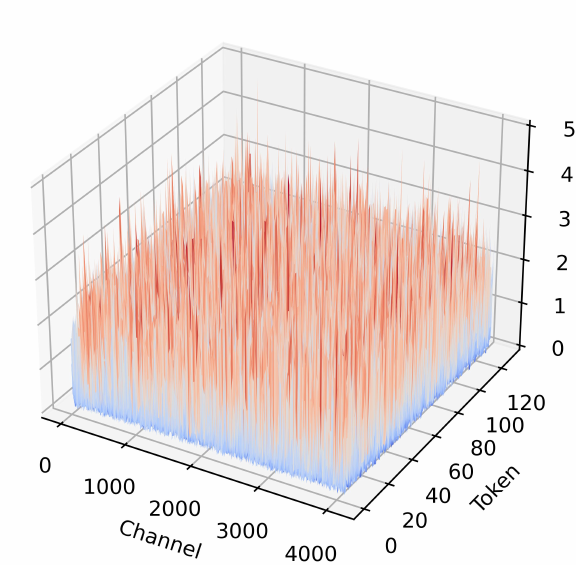}
  \caption{QuaRot}
  \label{fig:3}
\end{subfigure}
\hfill
\begin{subfigure}[b]{0.24\textwidth}
  \centering
  \includegraphics[width=\linewidth]{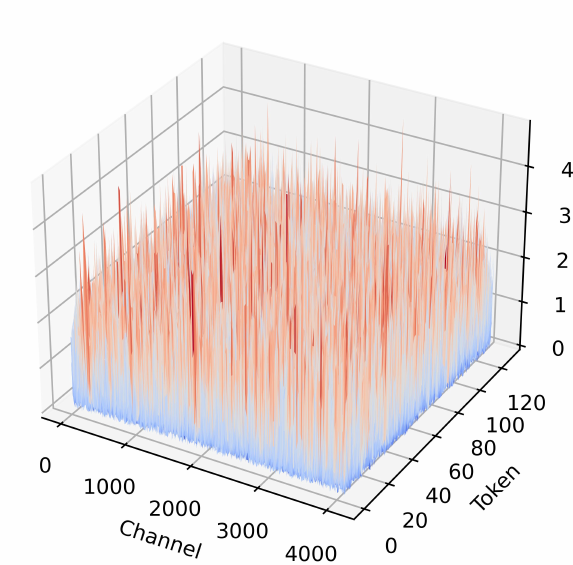}
  \caption{SpinQuant}
  \label{fig:4}
\end{subfigure}
\hfill
\begin{subfigure}[b]{0.24\textwidth}
  \centering
  \includegraphics[width=\linewidth]{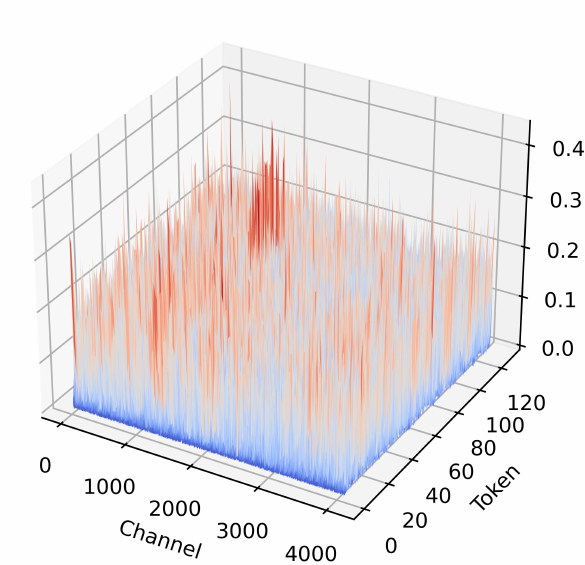}
  \caption{FlatQuant}
  \label{fig:5}
\end{subfigure}
\caption{Activation distributions of the \texttt{q\_proj} module in layer 8 of openPangu-Embedded-7B-V1.1 with different quantization methods. Each subplot shows the activations observed during inference, highlighting how quantization methods alter the dynamic range and distribution.}
\label{fig:activation_vis_layer8_qproj}
\end{figure*}

\subsection{Visualization of Activations}
Figure~\ref{fig:activation_vis_layer8_qproj} shows the activation distributions of the \texttt{q\_proj} module in Layer 8 of openPangu-Embedded-7B-V1.1 under various quantization methods.

\subsection{Use of AI Assistants}
We acknowledge that we used AI to help improve the manuscript, mainly for grammar, phrasing, and overall clarity. AI was also briefly used to fix small errors and syntax in the code included in the work.

%% file: tables/pangu7b_reasoning.tex
\begin{table*}[]
\centering
\begin{tabular}{llccccc}
\toprule[1.5pt]
\multirow{2}{*}{\textbf{Bits}} &
\multirow{2}{*}{\textbf{Method}} &
\multicolumn{3}{c}{\textbf{ACC (0-shot) ↑}} &
\multirow{2}{*}{\textit{\textbf{Avg.}}} &
\multirow{2}{*}{\textit{\textbf{Recovery (\%)}}} \\ 
\cmidrule(lr){3-5}
&
& \textbf{AIME24} & \textbf{AIME25} & \textbf{MATH-500} & & \\ 
\midrule

BF16                      & --               & 77.29 & 71.25 & 96.20 & 81.58 & 100.00 \\ 
\midrule
\multirow{5}{*}{W8A8}    & RTN              & 78.33 & 67.71 & 94.80 & 80.28 & \cellcolor{medium}{98.41} \\
                          & $\text{SpinQuant}^{*}$  & 76.88 & 68.96 & 96.20 & 80.68 & \cellcolor{best}{99.00} \\
                          & $\text{FlatQuant}^{*}$ &  74.58 & 65.62 & 94.20 & 78.13 & \cellcolor{worst}{95.78} \\
                          & $\text{SmoothQuant}^{*}$ & 77.29 & 67.71 & 95.40 & 80.13 & \cellcolor{medium}{98.23} \\
                          & GPTQ             &  80.00 & 68.54 & 95.00  & \textbf{81.18} & \cellcolor{best}{\textbf{99.51}} \\ 
\midrule

\multirow{5}{*}{W4A8}    & RTN              & 70.00 & 61.46 & 94.80 & 75.42 & \cellcolor{worst}{92.45} \\
                          & $\text{SpinQuant}^{*}$ & 73.96 & 66.46 & 94.00 & \textbf{78.14} & \cellcolor{worst}{\textbf{95.78}} \\
                          & $\text{FlatQuant}^{*}$ & 71.67 & 65.62 & 95.40  & 77.56 & \cellcolor{worst}{95.08} \\
                          & $\text{SmoothQuant}^{*}$ & 73.96 & 62.29 & 95.60  & 77.28 & \cellcolor{worst}{94.73} \\
                          & GPTQ             &  72.50 & 66.04 & 94.80 & 77.78 & \cellcolor{worst}{95.34} \\ 
\midrule

\multirow{5}{*}{W4A4}    & RTN              &  62.50 & 54.37 & 94.20 & 70.36 & \cellcolor{worst}{86.24} \\
                          & $\text{SpinQuant}^{*}$ &  47.26 & 44.37 & 90.80 & 60.81 & \cellcolor{worst}{74.54} \\
                          & $\text{FlatQuant}^{*}$ &  68.75 & 57.92 & 93.60 & \textbf{73.42} & \cellcolor{worst}{\textbf{90.00}} \\
                          & $\text{SmoothQuant}^{*}$ & 61.25 & 54.58 & 93.00 & 69.61 & \cellcolor{worst}{85.33} \\
                          & GPTQ             &  66.67 & 55.21 & 93.00 & 71.63 & \cellcolor{worst}{87.80} \\ 
\bottomrule[1.5pt]
\end{tabular}
\caption{Comprehensive comparison of PTQ methods on \textbf{openPangu-Embedded-7B-V1.1} under W8A8, W4A8, and W4A4 in terms of \textbf{reasoning} downstream task accuracy. * denotes the variant integrated with the GPTQ algorithm. Recovery (\%) is computed relative to the BF16 baseline.}
\label{pangu_math_reasoning}
\end{table*}

%% file: tables/pangu_7bvl.tex
\begin{table*}[!t]
\renewcommand{\arraystretch}{1.0}
\resizebox{\linewidth}{!}{
\begin{tabular}{llccccccc|c}
\toprule[1.5pt]
\textbf{Bits} & \textbf{Method} &
\textbf{OCRBench} &
\textbf{MMBench} &
\textbf{$\text{MMBench}^{CN}$} &
\textbf{TextVQA} &
\textbf{ChartQA} &
\textbf{MME} &
\textbf{MMMU} &
\textit{\textbf{Recovery~(\%)}}  \\
\cmidrule(lr){1-9} \cmidrule(l){10-10}
BF16 & --               & 918 & 85.46 & 85.4 & 83.92 & 87.68 & 2287 & 54.5 & 100.00 \\ 
\cmidrule{1-10} 

\multirow{7}{*}{W8A8} & RTN           & 906 & 85.05 & 85.71 & 83.78 & 88.04 & 2329 & 51.8 & \cellcolor{best}99.39  \\
     & QuaRot        & 899 & 85.54 & 84.54 & 83.64 & 87.32 & 2263 & 53.9 & \cellcolor{best}99.16 \\
     & QuaRot*       & 898 & 85.80 & 84.36 & 83.48 & 87.52 & 2266 & 52.4 & \cellcolor{medium}98.79 \\
     & SpinQuant     & 905 & 85.42 & 85.31 & 83.84 & 87.25 & 2285 & 54.2 & \cellcolor{best}99.60 \\
     & SpinQuant*    & 904 & 85.46 & 85.21 & 83.71 & 87.24 & 2285 & 54.3 & \cellcolor{best}99.58 \\
     & FlatQuant     & 908  & 85.20& 85.05 & 83.79& 87.42  & 2280 & 54.3 & \cellcolor{best}99.58 \\
     & FlatQuant*    & 908 & 85.46 & 85.65 & 84.14 & 87.56 & 2285 & 54.6 & \cellcolor{best}\textbf{99.92} \\
     & AWQ           & 911 & 85.22 & 85.20 & 83.76 & 87.68 & 2290 & 54.0   & \cellcolor{best}99.68 \\
     & SmoothQuant   & 914 & 85.29 & 85.48 & 82.71 & 87.66 & 2289 & 54.4 & \cellcolor{best}99.70 \\
     & SmoothQuant*  & 908 & 85.29 & 85.21 & 83.14 & 87.25 & 2280 & 54.5 & \cellcolor{best}99.54 \\
     & GPTQ          & 908 & 85.71 & 85.22 & 83.93 & 87.72 & 2295 & 54.3 & \cellcolor{best}{99.86} \\ 
\cmidrule{1-10} 

\multirow{7}{*}{W4A8} & RTN           & 887 & 83.08 & 82.90  & 82.47 & 86.52 & 2237 & 51.3 & \cellcolor{medium}97.11 \\
     & QuaRot        & 872 & 81.97 & 80.41 & 82.16 & 86.12 & 2270 & 47.7 & \cellcolor{worst}95.42 \\
     & QuaRot*       & 884 & 84.86 & 81.79 & 82.90 & 86.88 & 2269 & 50.7 & \cellcolor{medium}97.35 \\
     & SpinQuant     & 894 & 83.08 & 81.70 & 82.94 & 87.32 & 2231 & 49.6 & \cellcolor{worst}96.75 \\
     & SpinQuant*    & 911 & 84.10 & 83.08 & 83.65 & 86.80 & 2291 & 53.3 & \cellcolor{medium}98.80 \\
     & FlatQuant     & 906 & 84.18 & 83.42 & 83.44 & 87.52 & 2280 & 53.3 & \cellcolor{medium}98.80 \\
     & FlatQuant*    & 905 & 84.27 & 84.54 & 83.49 & 87.80 & 2259 & 53.3 & \cellcolor{medium}\textbf{98.91} \\
     & AWQ           & 884 & 83.59 & 82.65 & 82.80 & 86.48 & 2264 & 53.2 & \cellcolor{medium}97.82 \\
     & SmoothQuant   & 884 & 83.16 & 82.39 & 82.71 & 87.16 & 2259 & 52.7 & \cellcolor{medium}97.65 \\
     & SmoothQuant*  & 903 & 83.93 & 84.11 & 83.23 & 87.48 & 2289 & 51.3 & \cellcolor{medium}98.32 \\
     & GPTQ          & 898 & 85.63 & 84.19 & 83.11 & 87.24 & 2267 & 52.8 & \cellcolor{medium}98.73 \\ 
\cmidrule{1-10} 

\multirow{7}{*}{W4A4} & RTN           & 878 & 81.55 & 80.50  & 80.77 & 85.88 & 2193 & 52.3 & \cellcolor{worst}95.91 \\
     & QuaRot        & 823 & 68.11 & 53.61 & 76.26 & 68.10 & 2005 & 40.1 & \cellcolor{worst}80.27 \\
     & QuaRot*       & 844 & 74.40 & 62.63 & 77.97 & 78.84 & 2034 & 44.5 & \cellcolor{worst}86.54 \\
     & SpinQuant     & 852 & 72.45 & 63.06 & 78.56 & 83.40 & 1988 & 41.3 & \cellcolor{worst}86.12 \\
     & SpinQuant*    & 874 & 77.98 & 75.09 & 80.45 & 85.90 & 2198 & 48.3 & \cellcolor{worst}93.28 \\
     & FlatQuant     & 866 & 82.99 & 81.96 & 81.43 & 86.44 & 2191 & 50.0 & \cellcolor{worst}95.80  \\
     & FlatQuant*    & 896 & 84.01 & 81.7 & 83.15 & 86.96 & 2196 & 50.3 & \cellcolor{worst}\textbf{96.88}  \\
     & AWQ           & 880 & 81.80  & 80.67 & 81.04 & 85.80  & 2226 & 50.5 & \cellcolor{worst}95.78 \\
     & SmoothQuant   & 867 & 80.61 & 78.69 & 80.89 & 85.64 & 2215 & 52.4 & \cellcolor{worst}95.42  \\
     & SmoothQuant*  & 892 & 81.63 & 80.15 & 82.2 & 87.04 & 2246 & 50.6 & \cellcolor{worst}96.40  \\
     & GPTQ          & 879 & 82.56 & 80.58 & 81.00 & 85.88 & 2277 & 51.0 & \cellcolor{worst}{96.33}  \\ 
\bottomrule[1.5pt]
\end{tabular}
}
\caption{Comparison of PTQ methods on \textbf{openPangu-VL-7B} under W8A8, W4A8, and W4A4 quantization across multimodal benchmarks. * denotes the variant integrated with the GPTQ algorithm. 
}
\label{tab:pangu7bvl}
\end{table*}

%% file: tables/pangu7b_all.tex
\begin{table*}[]
\renewcommand{\arraystretch}{0.75}
\resizebox{\linewidth}{!}{
\begin{tabular}{llccccccc|c}
\toprule[1.5pt]
\multirow{2}{*}{\textbf{Bits}} &
\multirow{2}{*}{\textbf{Method}} &
\multicolumn{7}{c|}{\textbf{ACC (0-shot) ↑}} &
\textbf{PPL ↓} \\ \cmidrule(lr){3-9} \cmidrule(l){10-10}
&
&
\multicolumn{1}{l}{\textbf{ARC-C}} &
\multicolumn{1}{l}{\textbf{ARC-E}} &
\multicolumn{1}{l}{\textbf{HellaSwag}} &
\multicolumn{1}{l}{\textbf{PIQA}} &
\multicolumn{1}{l}{\textbf{Winogrande}} &
\textit{\textbf{Avg.}} &
\textit{\textbf{Recovery.(\%)}} &
\multicolumn{1}{l}{\textbf{WikiText}} \\ \midrule

BF16                      & --               & 42.75 & 67.26 & 62.86 & 73.50 & 60.62 & 61.40 & 100.00 & 34.89 \\ \cmidrule{1-10} 
\multirow{12}{*}{W8A8}    & RTN              & 44.03 & 66.25 & 61.74 & 73.61 & 57.77 & 60.68 & \cellcolor{medium}{98.83} & 35.75 \\
                          & QuaRot           & 42.24 & 65.95 & 62.45 & 72.42 & 60.69 & 60.75 & \cellcolor{medium}98.94 & 35.21 \\
                          & $\text{QuaRot}^{*}$           & 42.83 & 68.22 & 62.18 & 73.78 & 60.96 & 61.59 & \cellcolor{best}{100.32} & 35.75 \\
                          & SpinQuant        & 42.49 & 67.34 & 62.42 & 72.31 & 60.30 & 60.97 & \cellcolor{best}{99.31} & 35.49 \\
                          & $\text{SpinQuant}^{*}$  & 42.58 & 68.43 & 62.52 & 73.94 & 59.59 & 61.41 & \cellcolor{best}{100.02} & 33.04 \\
                          & FlatQuant        & 43.43 & 68.64 & 62.25 & 73.78 & 59.04 & 61.43 & \cellcolor{best}{100.05} & {30.96} \\
                          & $\text{FlatQuant}^{*}$ & 43.86 & 69.23 & 62.33 & 74.05 & 61.33 & \textbf{62.16} & \cellcolor{best}{\textbf{101.24}} & \textbf{30.87} \\
                          & AWQ              & 41.98 & 68.48 & 61.16 & 72.58 & 61.09 & 61.06 & \cellcolor{best}{99.45} & 38.00 \\
                          & SmoothQuant      & 42.58 & 66.50 & 62.08 & 72.69 & 59.19 & 60.61 & \cellcolor{medium}98.71 & 35.25 \\
                          & $\text{SmoothQuant}^{*}$ & 42.15 & 66.67 & 61.84 & 72.36 & 59.59 & 60.52 & \cellcolor{medium}{98.57} & {35.00} \\
                          & MR-GPTQ          & 43.09 & 67.80 & 62.84 & 73.72 & 59.91 & {61.47} & \cellcolor{best}{100.12}  & 34.57 \\
                          & GPTQ             & 42.92 & 67.13 & 61.66 & 72.69 & 57.85 & 60.45 & \cellcolor{medium}98.46 & 34.75 \\ \cmidrule{1-10} 
\multirow{12}{*}{W4A16}    & RTN              & 41.72 & 66.04 & 59.06 & 72.36 & 58.64 & 59.56 & \cellcolor{medium}97.01 & 39.99 \\
                          & QuaRot           & 40.53 & 63.05 & 58.25 & 71.76 & 57.70 & 58.26 & \cellcolor{worst}94.89 & 43.29 \\
                          & $\text{QuaRot}^{*}$ & 40.27 & 64.35 & 58.65 & 71.44 & 57.62 & 58.47 & \cellcolor{worst}95.22 & 40.50 \\
                          & SpinQuant        & 39.25 & 63.13 & 58.35 & 72.09 & 59.12 & 58.39 & \cellcolor{worst}95.10 & 38.11 \\
                          & $\text{SpinQuant}^{*}$ & 41.55 & 68.18 & 61.02 & 72.85 & 60.38 & \textbf{60.80} & \cellcolor{best}\textbf{99.02} & \textbf{35.26} \\
                          & FlatQuant        & 39.68 & 65.45 & 60.50 & 72.80 & 58.96 & 59.48 & \cellcolor{worst}96.87 & 37.68 \\
                          & $\text{FlatQuant}^{*}$ & 41.81 & 66.92 & 60.72 & 72.42 & 59.12 & 60.20 & \cellcolor{medium}98.05 & 37.62 \\
                          & AWQ              & 41.81 & 66.33 & 58.37 & 73.67 & 58.80 & 59.80 & \cellcolor{medium}97.39 & 40.00 \\
                          & SmoothQuant      & 40.44 & 66.08 & 59.71 & 72.80 & 57.22 & 59.25 & \cellcolor{worst}96.50 & 43.25 \\
                          & $\text{SmoothQuant}^{*}$ & 42.49 & 66.75 & 60.46 & 73.56 & 58.80 & 60.41 & \cellcolor{medium}98.39 & 38.75 \\

                          & MR-GPTQ          & -- & -- & -- & -- & -- & -- & -- & --                      \\
                          & GPTQ             & 40.80 & 65.50 & 60.80 & 72.20 & 59.75 & 59.81 & \cellcolor{medium}97.41 & {37.00}  \\ \cmidrule{1-10} 
\multirow{12}{*}{W4A8}    & RTN              & 39.59 & 65.24 & 58.70 & 71.60 & 57.85 & 58.60 & \cellcolor{worst}95.44 & 42.17 \\
                          & QuaRot           & 40.10 & 64.10 & 57.95 & 72.69 & 57.70 & 58.51 & \cellcolor{worst}95.29 & 40.87 \\
                          & $\text{QuaRot}^{*}$ & 40.02 & 65.07 & 58.54 & 71.44 & 57.62 & 58.54 & \cellcolor{worst}95.84 & 43.83 \\
                          & SpinQuant        & 39.59 & 61.57 & 58.56 & 72.09 & 58.17 & 58.00 & \cellcolor{worst}94.46 & 38.94 \\
                          & $\text{SpinQuant}^{*}$ & 42.41 & 67.85 & 60.74 & 72.91 & 58.8 & \textbf{60.54} & \cellcolor{medium}\textbf{98.60} & \textbf{34.75} \\
                          & FlatQuant        & 40.78 & 66.54 & 60.15 & 72.52 & 58.17 & 59.63 & \cellcolor{medium}97.12 & 37.67 \\
                          & $\text{FlatQuant}^{*}$ & 42.15 & 67.63 & 60.54 & 72.80 & 57.38 & 60.10 & \cellcolor{medium}97.89 & 37.72 \\
                          & AWQ              & 39.85 & 66.25 & 58.13 & 73.07 & 58.17 & 59.09 & \cellcolor{worst}96.25 & 41.75 \\
                          & SmoothQuant      & 40.78 & 66.20 & 58.82 & 72.47 & 57.46 & 59.15 & \cellcolor{worst}96.33 & 43.25 \\
                          & $\text{SmoothQuant}^{*}$ & 41.64 & 67.47 & 59.88 & 73.07 & 58.88 & 60.19 & \cellcolor{medium}98.03 & 40.00 \\

                          & MR-GPTQ          & 40.87 & 66.84 & 60.64 & 72.63 & 57.93 & {59.78} & \cellcolor{medium}{97.36} & 39.19 \\
                          & GPTQ             & 40.96 & 65.51 & 59.48 & 72.42 & 59.51 & 59.58 & \cellcolor{medium}97.03 & {37.50}  \\ \cmidrule{1-10} 
                          
\multirow{12}{*}{W4A8KV8} & RTN              & 40.19 & 63.34 & 57.94 & 72.03 & 58.56 & 58.41 & \cellcolor{worst}95.14 & 44.16 \\
                          & QuaRot           & 39.93 & 63.76 & 57.73 & 72.03 & 57.93 & 58.28 & \cellcolor{worst}94.92 & 41.74 \\
                          & $\text{QuaRot}^{*}$ & 39.93 & 64.69 & 58.13 & 71.06 & 59.27 & 58.62 & \cellcolor{worst}95.47 & 42.50 \\
                          & SpinQuant        & 37.63 & 62.96 & 58.06 & 72.14 & 58.80 & 57.92 & \cellcolor{worst}94.33 & 39.81 \\
                          & $\text{SpinQuant}^{*}$ & 43.17 & 65.78 & 60.34 & 72.36 & 60.69 & \textbf{60.47} & \cellcolor{medium}\textbf{98.48} & \textbf{36.56} \\
                          & FlatQuant        & 40.61 & 68.10 & 59.58 & 72.31 & 58.48 & 59.82 & \cellcolor{medium}97.42 & 38.66 \\
                          & $\text{FlatQuant}^{*}$ & 41.81 & 66.29 & 60.05 & 71.76 & 58.01 & 59.58 & \cellcolor{medium}97.03 & 39.28 \\
                          & AWQ              & 39.16 & 64.94 & 56.87 & 72.31 & 58.17 & 58.29 & \cellcolor{worst}94.94 & 44.50 \\
                          & SmoothQuant      & 39.85 & 64.60 & 58.20 & 71.87 & 56.99 & 58.30 & \cellcolor{worst}94.96 & 45.25 \\
                          & $\text{SmoothQuant}^{*}$ & 42.49 & 64.44 & 59.15 & 71.76 & 58.96 & 59.36 & \cellcolor{worst}96.68 & 41.25 \\

                          & MR-GPTQ          & -- & -- & -- & -- & -- & -- & -- & --                   \\
                          & GPTQ             & 39.51 & 63.41 & 58.89 & 71.00 & 59.83 & 58.53 & \cellcolor{worst}95.32 & {39.25}  \\ \cmidrule{1-10} 
                          
\multirow{12}{*}{W4A4}    & RTN              & 38.48 & 61.53 & 56.42 & 70.46 & 56.91 & 56.76 & \cellcolor{worst}92.45 & 49.33 \\
                          & QuaRot           & 36.43 & 56.19 & 51.24 & 66.76 & 54.54 & 53.03 & \cellcolor{worst}86.37 & 56.19 \\
                          & $\text{QuaRot}^{*}$ & 36.09 & 59.30 & 53.06 & 68.55 & 55.33 & 54.47& \cellcolor{worst}88.71 & 53.75 \\
                          & SpinQuant        & 34.64 & 57.24 & 52.82 & 68.39 & 55.49 & 53.72 & \cellcolor{worst}87.49 & 51.09 \\
                          & $\text{SpinQuant}^{*}$ & 37.88 & 60.4 & 55.34 & 69.53 & 57.22 & 56.07 & \cellcolor{worst}91.33 & 46.28 \\
                          & FlatQuant        & 39.93 & 65.82 & 57.07 & 71.00 & 56.04 & {57.97} & \cellcolor{worst}{94.42} & {38.36} \\
                          & $\text{FlatQuant}^{*}$ & 40.19 & 66.84 & 58.13 & 69.80 & 57.54 & \textbf{58.50} & \cellcolor{worst}\textbf{95.28} & \textbf{36.40} \\
                          & AWQ              & 37.08 & 63.76 & 54.49 & 71.06 & 57.70 & 56.82 & \cellcolor{worst}92.54 & 46.00 \\
                          & SmoothQuant      & 38.99 & 65.07 & 55.81 & 70.80 & 56.70 & 57.47 & \cellcolor{worst}93.60 & 52.00 \\
                          & $\text{SmoothQuant}^{*}$ & 38.65 & 64.60 & 56.74 & 70.73 & 57.62 & 57.67 & \cellcolor{worst}93.92 & 43.57 \\
                          & MR-GPTQ          & 38.99 & 63.43 & 57.19 & 69.48 & 58.96 & 57.61 & \cellcolor{worst}93.83 & 42.17 \\
                          & GPTQ             & 39.25 & 62.75 & 57.47 & 69.91 & 56.99 & 57.27 & \cellcolor{worst}93.28 & 44.50 \\ \bottomrule[1.5pt]
\end{tabular}
}
\caption{Comprehensive comparison of PTQ methods on \textbf{openPangu- 
Embedded-7B-V1.1} under W8A8, W4A8, and W4A4 in terms of \textbf{non-reasoning} downstream task accuracy and perplexity. * denotes the variant integrated with the GPTQ algorithm.}
\label{pangu_nonreasoning_all}
\end{table*}

%% file: tables/llama8b_all.tex
\begin{table*}[htbp]
\centering
\renewcommand{\arraystretch}{0.75}
\resizebox{\linewidth}{!}{
\begin{tabular}{llccccccc|c}
\toprule[1.5pt]
\multirow{2}{*}{\textbf{Bits}} &
\multirow{2}{*}{\textbf{Method}} &
\multicolumn{7}{c|}{\textbf{ACC (0-shot) ↑}} &
\textbf{PPL ↓} \\
\cmidrule(lr){3-9} \cmidrule(l){10-10}
&
& \textbf{ARC-C} & \textbf{ARC-E} & \textbf{HellaSwag} & \textbf{PIQA} & \textbf{Winogrande} & \textit{\textbf{Avg.}} & \textit{\textbf{Recovery (\%)}} & \textbf{WikiText} \\
\midrule

BF16 & -- & 55.20 & 79.63 & 79.15 & 81.07 & 73.95 & 73.80 & 100.00 & -- \\

\cmidrule{1-10}

\multirow{12}{*}{W8A8}
 & RTN               & 53.50 & 78.45 & 79.11 & 80.14 & 73.64 & 72.97 & \cellcolor{medium}98.87 & 7.31 \\
 & QuaRot            & 55.72 & 80.09 & 78.76 & 80.58 & 73.64 & 73.76 & \cellcolor{best}99.94 & 7.39 \\
 & $\text{QuaRot}^{*}$            & 56.4 & 80.81 & 78.84 & 81.01 & 74.35 & \textbf{74.28} & \cellcolor{best}\textbf{100.65} & 7.40 \\
 & SpinQuant         & 55.89 & 80.18 & 78.60 & 80.79 & 74.51 & 73.99 & \cellcolor{best}100.26 & 7.39 \\
 & $\text{SpinQuant}^{*}$            & 54.35 & 80.09 & 78.12 & 80.2 & 73.8 & 73.31 & \cellcolor{best}99.34 & 7.41 \\
 & FlatQuant         & 55.80 & 79.46 & 78.99 & 81.12 & 73.80 & 73.83 & \cellcolor{best}100.05 & 7.27 \\
 & $\text{FlatQuant}^{*}$            & 54.69 & 79.42 & 78.56 & 80.63 & 74.19 & 73.50 & \cellcolor{best}99.59 & 7.33 \\
 & AWQ               & 54.27 & 79.17 & 78.48 & 80.74 & 73.80 & 73.29 & \cellcolor{best}99.31 & 7.34 \\
 & SmoothQuant       & 55.38 & 79.08 & 79.14 & 81.23 & 74.51 & 73.87 & \cellcolor{best}100.09 & 7.34 \\
 & $\text{SmoothQuant}^{*}$            & 55.03 & 79.63 & 78.73 & 81.39 & 73.88 & 73.73 & \cellcolor{best}99.91 & 7.34 \\
 & MR-GPTQ           & 55.38 & 79.12 & 78.99 & 81.01 & 73.95 & 73.69 & \cellcolor{best}99.85 & \textbf{7.24} \\
 & GPTQ              & 54.69 & 79.12 & 78.47 & 80.79 & 74.35 & 73.48 & \cellcolor{best}99.57 & 7.33 \\

\cmidrule{1-10}

\multirow{12}{*}{W4A16}
 & RTN               & 53.24 & 77.90 & 77.32 & 80.36 & 73.80 & 72.52 & \cellcolor{medium}98.27 & 7.64 \\
 & QuaRot            & 48.98 & 75.67 & 76.68 & 78.56 & 70.64 & 70.11 & \cellcolor{worst}94.99 & 8.28 \\
 & $\text{QuaRot}^{*}$ & 53.33 & 77.65 & 77.00 & 79.43 & 73.40 & 72.16 & \cellcolor{medium}97.78 & 7.75 \\
 & SpinQuant         & 50.43 & 75.13 & 76.17 & 78.67 & 71.51 & 70.38 & \cellcolor{worst}95.37 & 8.38 \\
 & $\text{SpinQuant}^{*}$            & 52.47 & 78.37 & 77.35 & 80.2 & 73.4 & 72.36 & \cellcolor{medium}98.05 & 7.70 \\
 & FlatQuant         & 53.92 & 77.36 & 77.58 & 80.25 & 73.32 & 72.49 & \cellcolor{medium}98.22 & 7.86 \\
 & $\text{FlatQuant}^{*}$  & -- & -- & -- & -- & -- & -- & -- & -- \\
 & AWQ               & 53.92 & 79.46 & 77.73 & 80.25 & 73.40 & {72.95} & \cellcolor{medium}{98.85} & \textbf{7.62} \\
 & SmoothQuant       & 53.50 & 78.48 & 77.85 & 80.25 & 74.51 & 72.92 & \cellcolor{medium}98.80 & 7.62 \\
 & $\text{SmoothQuant}^{*}$            & 53.16 & 76.81 & 77.61 & 79.11 & 73.01 & 71.94 & \cellcolor{medium}97.48 & 7.75 \\
 & MR-GPTQ           & -- & -- & -- & -- & -- & -- & -- & -- \\
 & GPTQ              & 53.92 & 79.04 & 77.65 & 80.96 & 73.48 & \textbf{73.01} & \cellcolor{medium}\textbf{98.93} & 7.71 \\

\cmidrule{1-10}

\multirow{12}{*}{W4A8}
 & RTN               & 53.07 & 76.81 & 77.04 & 80.03 & 73.48 & 72.09 & \cellcolor{medium}97.68 & \textbf{7.71} \\
 & QuaRot            & 49.15 & 75.93 & 76.37 & 78.29 & 71.03 & 70.15 & \cellcolor{worst}95.06 & 8.45 \\
 & $\text{QuaRot}^{*}$            & 52.65 & 79.67 & 77.86 & 80.52 & 73.40 & 72.82 & \cellcolor{medium}98.67 & 7.95 \\
 & SpinQuant         & 49.91 & 75.59 & 76.10 & 78.07 & 71.90 & 70.31 & \cellcolor{worst}95.28 & 8.50 \\
 & $\text{SpinQuant}^{*}$            & 52.82 & 78.37 & 76.96 & 79.6 & 73.32 & 72.21& \cellcolor{medium}97.85 & 7.84 \\
 & FlatQuant         & 53.84 & 80.26 & 77.27 & 80.03 & 71.82 & 72.64 & \cellcolor{medium}98.43 & 7.87 \\
 & $\text{FlatQuant}^{*}$            & 53.84 & 80.26 & 77.27 & 80.03 & 71.82 & 72.64 & \cellcolor{medium}97.95 & 7.81 \\
 & AWQ               & 53.84 & 79.25 & 77.55 & 79.98 & 73.88 & 72.90  & \cellcolor{medium}98.78 & 7.75 \\
 & SmoothQuant       & 54.01 & 77.61 & 77.77 & 79.98 & 74.98 & 72.87  & \cellcolor{medium}98.74 & 7.75 \\
 & $\text{SmoothQuant}^{*}$            & 52.96 & 77.02 & 77.42 & 78.63 & 70.64 & 71.33 & \cellcolor{worst}96.67 &  8.70\\
 & MR-GPTQ           & 54.52    & 78.62    & 77.28    & 80.47    & 73.80    & 72.94    & \cellcolor{medium}98.83 & 7.83 \\
 & GPTQ              & 53.5 & 79.59 & 79.71 & 81.18 & 73.6 & \textbf{73.52} & \cellcolor{best}\textbf{99.62} & 7.87 \\

\cmidrule{1-10}

\multirow{12}{*}{W4A8KV8}
 & RTN               & 52.22 & 77.02 & 76.91 & 79.92 & 73.64 & 71.94 & \cellcolor{medium}97.48 & \textbf{7.31} \\
 & QuaRot            & 50.09 & 76.09 & 76.25 & 78.78 & 70.32 & 70.31 & \cellcolor{worst}95.27 & 8.46 \\
 & $\text{QuaRot}^{*}$            & 52.73 & 79.92 & 77.78 & 80.03 & 73.40 & 72.77 & \cellcolor{medium}98.61 & 7.97 \\
 & SpinQuant         & 50.09 & 75.25 & 75.89 & 77.91 & 71.43 & 70.31 & \cellcolor{worst}95.28 & 8.50 \\
 & $\text{SpinQuant}^{*}$            & 52.9 & 78.32 & 77.44 & 80.58 & 72.61 & 72.37 & \cellcolor{medium}98.06 & 7.79 \\
 & FlatQuant         & 54.86 & 79.55 & 77.39 & 78.94 & 71.98 & 72.54 & \cellcolor{medium}98.30 & 7.87 \\
 & $\text{FlatQuant}^{*}$            & 53.84 & 78.83 & 77.70 & 79.71 & 73.09 & 72.63 & \cellcolor{medium}98.42 & 7.72 \\
 & AWQ               & 54.52 & 79.12 & 77.53 & 79.87 & 72.93 & \textbf{72.79} & \cellcolor{medium}\textbf{98.64} & 7.75 \\
 & SmoothQuant       & 53.33 & 77.48 & 77.57 & 80.30 & 74.98 & 72.73 & \cellcolor{medium}98.55 & 7.73 \\
 & $\text{SmoothQuant}^{*}$            & 51.54 & 79.17 & 77.54 & 79.11 & 72.77 & 72.03 & \cellcolor{medium}97.60 & 7.87 \\
 & MR-GPTQ           & -- & -- & -- & -- & -- & -- & -- & -- \\
 & GPTQ              & 54.15 & 78.62 & 77.37 & 80.09 & 73.35 & 72.72 & \cellcolor{medium}98.53 & 7.34 \\

\cmidrule{1-10}

\multirow{12}{*}{W4A4}
 & RTN               & 49.66 & 75.80 & 75.48 & 79.05 & 70.48 & 70.09 & \cellcolor{worst}94.98 & 8.27 \\
 & QuaRot            & 44.11 & 71.63 & 71.82 & 75.14 & 64.88 & 65.52 & \cellcolor{worst}88.78 & 10.34 \\
 & $\text{QuaRot}^{*}$            & 49.15 & 74.62 & 74.26 & 77.48 & 68.98 & 68.90 & \cellcolor{worst}93.36 & 9.12 \\
 & SpinQuant         & 43.09 & 67.26 & 70.71 & 74.92 & 65.98 & 64.39 & \cellcolor{worst}87.25 & 10.40 \\
 & $\text{SpinQuant}^{*}$            & 49.15 & 74.54 & 74.38 & 76.99 & 69.53 & 68.92 & \cellcolor{worst}93.38 & 9.01 \\
 & FlatQuant         & 52.05 & 77.27 & 76.89 & 79.49 & 70.64 & 71.27 & \cellcolor{worst}96.57 & \textbf{8.03}\\
 & $\text{FlatQuant}^{*}$            & 51.19 & 78.58 & 76.77 & 78.94 & 71.67 & \textbf{71.43} & \cellcolor{worst}\textbf{96.79} & 8.06 \\
 & AWQ               & 52.30 & 77.95 & 76.13 & 78.92 & 69.46 & 70.95  & \cellcolor{worst}96.14 & 8.25 \\
 & SmoothQuant       & 51.37 & 76.52 & 76.12 & 79.00 & 72.38 & 71.08  & \cellcolor{worst}96.31 & 8.25 \\
 & $\text{SmoothQuant}^{*}$            & 50.51 & 73.36 & 75.27 & 76.82 & 67.17 & 68.63 & \cellcolor{worst}92.98 & 8.50 \\
 & MR-GPTQ           & 50.85 & 75.46 & 76.02 & 79.82 & 70.80 & 70.59 & \cellcolor{worst}95.65 & 8.34 \\
 & GPTQ              & 50.68 & 77.65 & 75.66 & 78.18 & 70.24 & 70.48 & \cellcolor{worst}95.50 & 8.37 \\

\bottomrule[1.5pt]
\end{tabular}
}
\caption{Comprehensive comparison of PTQ methods on \textbf{Llama-3.1-8B-Instruct} under W8A8, W4A8, and W4A4 in terms of \textbf{non-reasoning} downstream task accuracy and perplexity. * denotes the variant integrated with the GPTQ algorithm.}
\label{llama_nonreasoning_all}
\end{table*}